\documentclass[letterpaper]{article} 
\usepackage{aaai2027}  
\nocopyright
\usepackage[hyphens]{url}  
\usepackage{graphicx} 
\usepackage{natbib}  
\usepackage{caption} 
\usepackage{algorithm}
\usepackage{algorithmic}
\usepackage{amsmath}

\usepackage{microtype}
\usepackage{graphicx}
\usepackage{booktabs} 
\usepackage{multirow}
\usepackage[table,xcdraw]{xcolor}
\usepackage{arydshln}  
\usepackage{pifont}
\usepackage[table,xcdraw]{xcolor}

\usepackage{caption}
\usepackage{xcolor}

\usepackage{amsmath}
\usepackage{amssymb}
\usepackage{mathtools}
\usepackage{amsthm}
\usepackage{subcaption} 
\usepackage{arydshln}

\usepackage{newfloat}
\usepackage{listings}
\DeclareCaptionStyle{ruled}{labelfont=normalfont,labelsep=colon,strut=off} 
\floatstyle{ruled}
\newfloat{listing}{tb}{lst}{}
\floatname{listing}{Listing}

\usepackage{booktabs}

\title{E$^3$mo-Bench: A Scalable Benchmark for Multimodal Evoked and Expressed Emotion Understanding via Bayesian Pairwise Alignment}
\author {
    Lancheng Gao\textsuperscript{\rm 1},
    Ziheng Jia\textsuperscript{\rm 1},
    Shengyan Li\textsuperscript{\rm 1},
    Zixuan Xing\textsuperscript{\rm 1}, \\
    Jiarui Wang\textsuperscript{\rm 1},
    Huiyu Duan\textsuperscript{\rm 1},
    Xiongkuo Min\textsuperscript{\rm 1}\corresponding
}
\affiliations {
    \textsuperscript{\rm 1}Shanghai Jiao Tong University
}

\begin{document}

\maketitle

\begin{abstract}
Understanding both expressed and evoked emotions is critical for multimodal large language models (MLLMs) to achieve comprehensive affect-aware interactions. However, existing benchmarks typically examine expressed and evoked emotions in isolation or are constrained to coarse-grained and incomplete affective characterizations. To bridge this gap, we introduce \textbf{E$^3$mo-Bench}, a scalable benchmark comprising $12{,}314$ question-answer pairs across $2{,}524$ videos with predefined affective perspectives. It evaluates \underline{e}voked and \underline{e}xpressed \underline{emo}tion understanding via $3$ complementary tasks: emotion perception, open-vocabulary recognition, and valence-arousal-dominance (VAD) assessment. To efficiently scale reliable continuous annotations, we propose Bayesian Pairwise Alignment, which aggregates sparse, low-burden pairwise judgments into anchor-referenced VAD estimates. Furthermore, we develop \textbf{E$^3$mo-Score}, a training-free agent that aggregates complementary judgments from a five-model committee to improve VAD estimation. Extensive experiments validate the effectiveness of our framework and expose a pronounced performance skew between evoked and expressed emotion paradigms. These findings, coupled with MLLMs' persistent deficits in fine-grained recognition and dimensional assessment, chart a clear course for advancing multimodal emotional intelligence. 
\end{abstract}



\section{Introduction}  

\begin{figure*}[t] 
    \centering
    \includegraphics[width=0.95\textwidth]{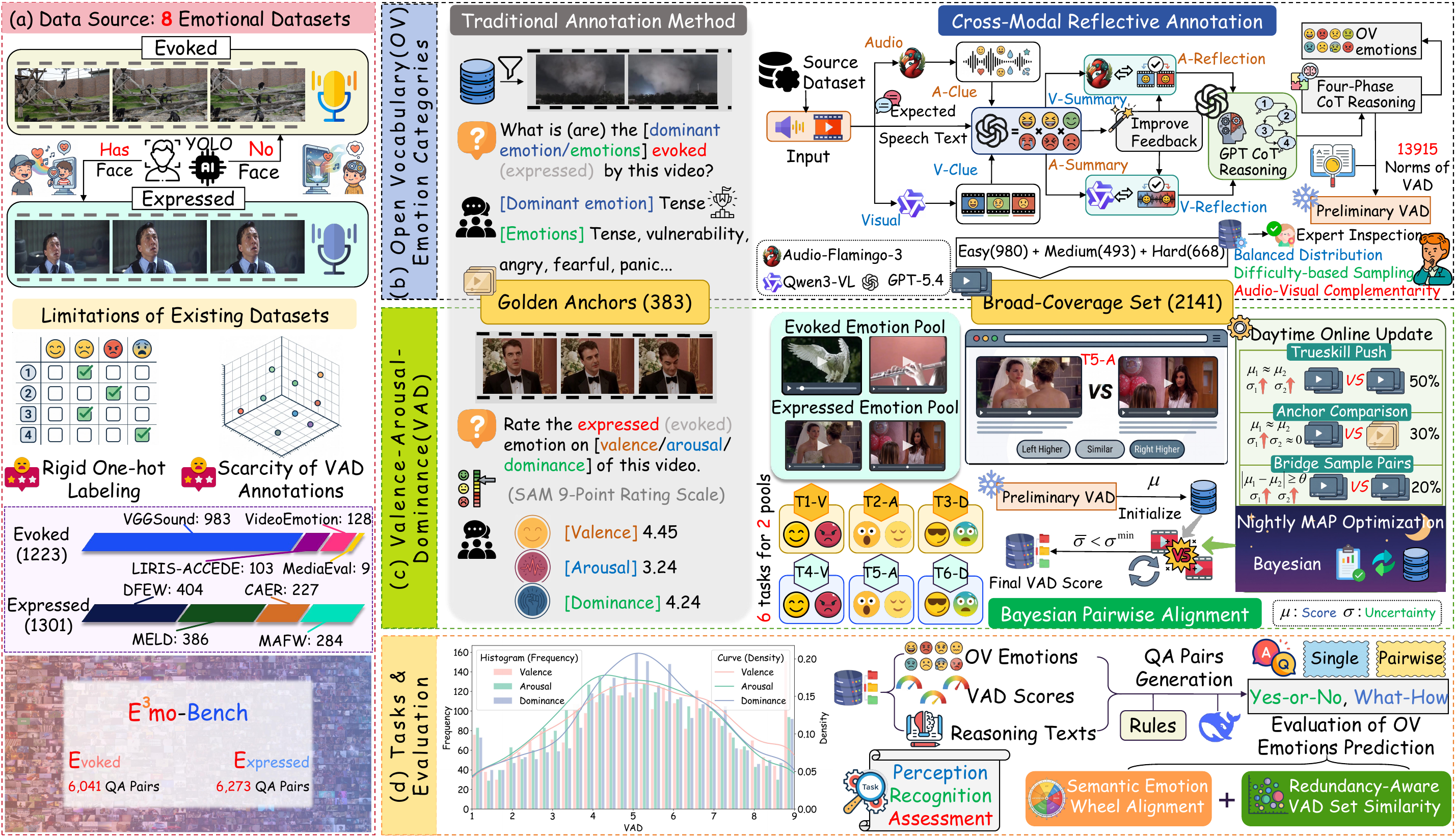}
    \caption{\textbf{E$^3$mo-Bench} construction pipeline. (a) Curation of fine-grained expressed and evoked emotion subsets from $8$ datasets. (b-c) Two-tier annotation schemes for OV emotion and VAD. (d) Generation of label-driven QA pairs covering $3$ tasks. }
    \label{fig: dataset construction}
    \vspace{-0.5cm}
\end{figure*}

Emotional intelligence (EI), defined as the capacity to perceive, interpret, and leverage emotional information~\cite{EI}, is increasingly critical for multimodal large language models (MLLMs) deployed in human-computer interaction~\cite{human-robot}, social robotics~\cite{human-robot-interaction}, and personalized media understanding~\cite{advertising}. A prerequisite for such intelligence is comprehensive multimodal emotion understanding, which requires bridging two complementary perspectives: subject-centered emotions expressed through facial and behavioral cues, and viewer-centered affective responses evoked by the broader audio-visual context. Assessing both perspectives is therefore essential for evaluating whether MLLMs exhibit generalizable EI beyond isolated, task-specific understanding.

Traditional audio-visual datasets typically isolate expressed and evoked emotions, collapsing samples into single closed-set categories~\cite{MELD, DFEW, VideoEmotion} and neglecting compound semantics alongside continuous valence-arousal-dominance (VAD) dimensions~\cite{VAD}. Although recent benchmarks have introduced open-vocabulary (OV) descriptions and causal reasoning to capture richer semantics, these advancements remain largely confined to expression-centric tasks~\cite{OV-MER,emobench-m}. Consequently, the fine-grained evaluation of evoked emotions in audio-visual contexts remains underexplored. A comprehensive assessment of EI necessitates an equally rigorous investigation of both paradigms, thereby demanding a unified benchmark that integrates fine-grained categorical and dimensional characterizations across both expressed and evoked perspectives.

A further obstacle lies in the scalability of audio-visual emotion annotation. OV annotation typically relies on humans to identify emotion concepts and explain their causes~\cite{MAFW}. However, annotators often focus on the most salient cues, easily leading to incomplete descriptions~\cite{AffectGPT}. For dimensional annotation, aggregating absolute ratings from multiple psychology-trained experts can reduce subjective bias but is costly and cognitively demanding at scale, particularly for abstract dimensions such as dominance. Relative judgments can lower this burden and often improve consistency~\cite{pagan}, but sparse comparisons do not directly yield globally calibrated scores and must account for ties, annotator reliability, and uneven coverage. This necessitates a scalable framework that synthesizes multimodal emotional evidence and recovers robust scores from efficient, low-burden human supervision.

To this end, we introduce \textbf{E$^3$mo-Bench}, a scalable benchmark designed for comprehensive multimodal evoked and expressed emotion understanding. Utilizing Cross-Modal Reflective Annotation (CMRA) across $8$ source datasets, we curate a $383$-sample Golden Anchor Set via strict manual annotation, alongside a $2{,}141$-sample Broad-Coverage Set featuring human-in-the-loop validation. Spanning diverse affective states and difficulty levels, this pipeline yields $12{,}314$ question-answer (QA) pairs distributed across $3$ core tasks: \textbf{Perception}, \textbf{Recognition}, and \textbf{Assessment}. To reliably scale continuous dimensional annotations, we propose Bayesian Pairwise Alignment (BPA), which adaptively schedules informative comparisons and periodically refines single-dimension scores via Bayesian maximum a posteriori (MAP) estimation. Building upon BPA, we deploy \textbf{E$^3$mo-Score}, a training-free agent that aggregates pairwise judgments from a five-model committee for robust VAD prediction. Extensive evaluations validate the effectiveness of our framework while exposing critical limitations in current MLLMs, especially their inability to achieve balanced, fine-grained understanding across affective perspectives. Our core contributions are threefold:

\begin{itemize}
\item We introduce \textbf{E$^3$mo-Bench}, a unified audio-visual benchmark that jointly evaluates evoked and expressed emotion understanding under a shared taxonomy of perception, OV emotion recognition, and VAD assessment. Each video is assigned one predefined affective perspective, yielding $12{,}314$ QA pairs across $2{,}524$ videos.

\item We propose \textbf{BPA}, a scalable annotation framework that derives anchor-calibrated scores from sparse pairwise judgments via informative comparison scheduling, periodic Bayesian optimization, and uncertainty estimation. Building upon this framework, we develop a training-free \textbf{E$^3$mo-Score} agent to automate robust VAD assessment.

\item Extensive experiments validate the effectiveness of E$^3$mo-Score and reveal pronounced imbalances between expressed and evoked emotion understanding, alongside weaknesses in fine-grained recognition and VAD assessment, providing crucial insights for future research.

\end{itemize}

\section{Related Works}  

\subsection{Audio-Visual Emotion Understanding Benchmarks}

Early audio-visual benchmarks isolate expressed and evoked emotions into discrete categories~\cite{DFEW, VideoEmotion} or continuous dimensions~\cite{LIRIS-ACCEDE, MediaEval}. Recent works advance finer-grained evaluations: OV-MER~\cite{OV-MER} and MME-Emotion~\cite{mme-emotion} introduce OV semantics and reasoning for expressed emotions, while EEmo-Bench~\cite{eemo-bench} and VAD Danmu~\cite{vad-danmu} capture evoked emotion through joint categorical-dimensional annotations. However, these benchmarks primarily emphasize one affective perspective. Although AICA-Bench~\cite{AICA-Bench} evaluates both perspectives, it is limited to static images and predefined categorical emotion spaces. \textbf{E$^3$mo-Bench} extends this dual-perspective setting to audio-visual content with OV emotion recognition and continuous VAD assessment.

\subsection{Scalable Affect Annotation and Pairwise Inference}

Recent studies have explored efficient paradigms for scaling affective datasets. For OV labeling, approaches like OV-MER~\cite{OV-MER} and EmoReAlM~\cite{AVERE} leverage MLLMs to extract and aggregate multimodal cues. However, bypassing validation of these intermediate cues often discards critical fine-grained semantics. For continuous affect, absolute VAD scoring via mean opinion
score~(MOS) \cite{eemo-bench} imposes a heavy cognitive load and suffers from subjective scale misalignment. To alleviate subjective biases, recent works increasingly pivot to relative judgment paradigms. These range from pairwise comparison platforms~\cite{pagan} and win-rate aggregations over random quadruplets~\cite{text_compare}, to more structured anchor-based derivations via Thurstone optimization~\cite{compare2score, thurstone}. However, translating these sparse relative comparisons into globally calibrated continuous scores intrinsically demands highly sophisticated task routing and rigorous optimization algorithms. To this end, we implement a \textbf{CMRA} pipeline for scalable OV labeling, and propose the \textbf{BPA} framework (featuring online routing and Bayesian optimization) alongside the E$^3$mo-Score agent to automate robust VAD rating.

\section{Benchmark Construction}

E$^3$mo-Bench represents expressed and evoked emotion through two complementary spaces (OV emotions and VAD), employing two distinct annotation schemes to label all data dimensions. The construction pipeline is illustrated in Fig.~\ref{fig: dataset construction}.

\subsection{Video Collection and Candidate Curation}

E$^3$mo-Bench is built upon $8$ diverse audio-visual datasets. We adopt DFEW~\cite{DFEW}, MELD~\cite{MELD}, CAER~\cite{CAER}, and MAFW~\cite{MAFW} to model expressed emotions. To mitigate the scarcity of evoked emotion resources, we combine LIRIS-ACCEDE~\cite{LIRIS-ACCEDE}, VideoEmotion~\cite{VideoEmotion}, and MediaEval~\cite{MediaEval} with the general-purpose VGGSound~\cite{VGGSound} as a supplementary source. Since these datasets lack OV labels and fine-grained 9-point SAM VAD scores~\cite{SAM}, they serve as an ideal foundation for our rigorous re-annotation pipeline.

\textbf{To evaluate diverse affective scenarios under a consistent protocol, we assign each video one primary affective perspective.} Although expressed and evoked emotions may coexist, this assignment specifies the intended evaluation target and the dominant affective evidence to be interpreted. Expressed-emotion instances focus on emotions conveyed by human subjects through facial and bodily cues, whereas evoked-emotion instances focus on affective responses associated with the broader scene and audio-visual context. Accordingly, face visibility naturally emerges as an operational cue for candidate routing. We employ \textit{YOLO11n}~\cite{yolo11} for face-based screening and retain only samples with unambiguous detection outcomes for subset assignment. This procedure limits target ambiguity and yields complementary subsets for perspective-specific evaluation.

\subsection{Multimodal Emotion Evidence Bootstrapping}

For sample validation, we generate preliminary OV labels and rationales via a model-based pipeline for reliability-stratified filtering. While prior methods like AffectGPT~\cite{AffectGPT} extract cues via isolated unimodal models and fuse them using a proprietary large language model, this single aggregation step cannot fully correct the biased attributions stemming from the initial modality isolation. To address this, we propose CMRA, a two-round reflective annotation framework based on cross-modal feedback and reanalysis.
Firstly, \textit{Qwen3-VL}~\cite{qwen3} and \textit{Audio-Flamingo-3}~\cite{audio-flamingo} extract visual and audio cues. \textit{GPT-5.4}~\cite{gpt5.4} then performs a preliminary fusion to identify conflicts and generate modality-specific revision suggestions. By feeding this complementary context back to the original models, they refine their analyses based on their observable evidence. Finally, \textit{GPT-5.4} synthesizes these updated cues into an OV emotion list with structured rationales, effectively resolving unimodal blind spots at the source.
Based on these rationales, we derive preliminary VAD scores by mapping extracted keywords to normative ratings of $13{,}915$ English lemmas~\cite{VAD_norm}, establishing crucial criteria for subsequent benchmark sampling.

\subsection{Golden Anchor Set Construction}

 Since emotion assessment is inherently subjective and cognitively demanding, we aggregate multiple annotations to neutralize scaling biases. Specifically, we stratify $200$ representative samples from each of the evoked and expressed subsets based on preliminary VAD scores. Under the protocol of ITU~\cite{ITU-R}, $15$ volunteers annotate each sample with OV labels and VAD ratings. To normalize subjective bias, these scores undergo intra-annotator $z$-score standardization before being rescaled to the $1$–-$9$ range. Ultimately, $383$ high-consensus samples are retained. Finally, OV labels are filtered via vote concentration and semantic centrality, while \textit{GPT-5.4} synthesizes the retained labels and preliminary affective cues into coherent Chain-of-Thought (CoT) rationales. Details on inter-rater reliability assessment and result aggregation are provided in Supplementary Material (\textit{Supp.}).


\subsection{Broad-Coverage Set Construction}

Traditional human annotation faces two critical bottlenecks: 1) Absolute VAD scoring demands rigorous scale calibration; otherwise, annotator-specific scale shifts and cognitive fatigue compromise accuracy. 2) Deriving reliable MOS requires numerous annotators per sample, severely restricting scalability. To overcome these hurdles, we design a scalable pipeline to construct the Broad-Coverage Set. Candidates are initially filtered using model-generated bootstrap annotations. To quantify affective complexity, each sample must contain sufficient OV terms mappable to a VAD lexicon. We then apply Difficulty-Stratified Sample Selection to categorize samples into $3$ tiers: 1) easy ($\sim 50\%$), exhibiting consistent cross-modal affective polarity and concentrated OV distributions (detailed in \textit{Supp.}); 2) hard ($\sim 30\%$), featuring pronounced cross-modal conflicts or mixed polarities (e.g., coexisting joy and sorrow), retained only if human-verified for contextual plausibility to test mixed-emotion understanding; and 3) medium ($\sim 20\%$), representing moderate affective dispersion between clear and complex cases while ensuring an overall balanced VAD distribution. After manually verifying all OV labels and CoT rationales and deriving the final VAD scores via \textbf{BPA (detailed in the next section)}, we ultimately secure $1{,}033$ evoked-emotion and $1{,}108$ expressed-emotion videos.

\begin{figure*}[t] 
    \centering
    \includegraphics[width=0.95\textwidth]{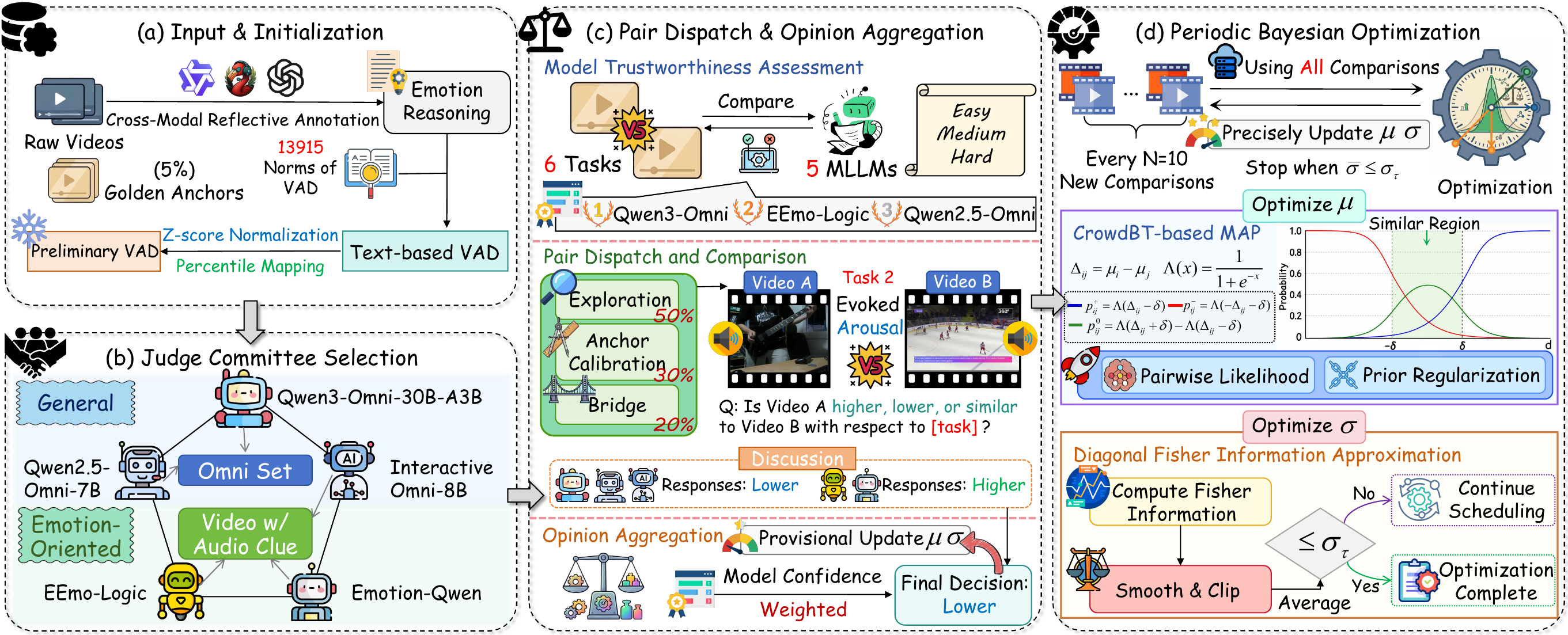}
    \caption{Overview of \textbf{E$^3$mo-Score}, a model-based instantiation of \textbf{BPA}: (a) sample initialization; (b) committee construction and calibration; (c) pair scheduling and judgment aggregation; and (d) periodic Bayesian refinement of scores and uncertainties.}
    \label{fig: BPA agent}
    \vspace{-0.5cm}
\end{figure*}

\subsection{Benchmark Task Generation}

Based on the human-annotated emotion labels, $3$ tasks are defined: Perception, Recognition, and Assessment.

\subsubsection{Perception.} 
This task encompasses both single-video and video-pair settings, featuring questions in \textit{Yes-or-No} and \textit{What-How} formats. To ensure balanced answer distribution, QA pairs are systematically synthesized using predefined rules anchored in OV labels and VAD scores. For inquiries regarding affective causes, \textit{Deepseek-v4-pro}~\cite{deepseek-v4} derives candidate options from existing rationales, followed by rigorous human verification.

\subsubsection{Recognition.} 
This task assesses fine-grained expressed and evoked emotion recognition. While we adopt the Semantic Emotion Wheel Alignment protocol ($F^{\text{EW}}$)~\cite{OV-MER} for basic consistency, it captures only semantic proximity and ignores error severity (e.g., confusing \textit{Surprise} with \textit{Joy} warrants a smaller penalty than confusing \textit{Anger} with \textit{Joy}). To address this, we introduce \textbf{Redundancy-Aware VAD Set Similarity}. By comparing predicted and reference sets within the continuous VAD space, this metric provides a nuanced evaluation of cross-cluster errors. It is defined as follows:

Let $G_i$ and $P_i$ denote the reference and predicted OV emotion sets for the $i$-th sample. To mitigate intra-set semantic redundancy, we map each emotion to its lexicon VAD scores, followed by $z$-score normalization to remove scale disparities. We then define the affective distance $d(e_i, e_j)$ between any two emotions via their Euclidean distance in this normalized VAD space. Emotions satisfying $d(e_i, e_j) \leq \tau_{\mathrm{dup}}$ are aggregated into local affective clusters, retaining only the emotion closest to the cluster centroid as the representative. This deduplication process yields the refined sets $G^{*}_i$ and $P^{*}_i$. For emotions across distinct semantic clusters, we define their pairwise VAD similarity $S(e_i, e_j)$ using a Gaussian kernel:

\vspace{-0.2cm}
\begin{equation}
    S(e_i,e_j)=\exp\left(
-\frac{ d(e_i,e_j)^2}{2\sigma_s^2}
\right)^{\gamma},
\end{equation}

\noindent where $\sigma_s$ dictates the similarity bandwidth and $\gamma$ calibrates sensitivity to subtle affective deviations. To compute the final score, we employ a bidirectional soft-coverage scheme that averages each emotion’s maximum similarity to the opposite set in both directions, formally defined as follows:


\vspace{-0.1cm}
\begin{equation}
C_i^{G\rightarrow P}
=
\frac{1}{\left|G_i^{*}\right|}
\sum_{g\in G_i^{*}}
\max_{p\in P_i^{*}} S(g,p),
\label{eq:vad_recall}
\end{equation}

\vspace{-0.1cm}
\begin{equation}
C_i^{P\rightarrow G}
=
\frac{1}{\left|P_i^{*}\right|}
\sum_{p\in P_i^{*}}
\max_{g\in G_i^{*}} S(p,g),
\label{eq:vad_precision}
\end{equation}

\vspace{-0.1cm}
\begin{equation}
\begin{aligned}
F_i^{\mathrm{VAD}}
=
\frac{ 2C_i^{G\rightarrow P}C_i^{P\rightarrow G} }{ C_i^{G\rightarrow P}+C_i^{P\rightarrow G} }
\left(
\frac{
\min\!\left(\left|G_i^{*}\right|,\left|P_i^{*}\right|\right)
}{
\max\!\left(\left|G_i^{*}\right|,\left|P_i^{*}\right|\right)
}
\right)^{\beta}.
\end{aligned}
\label{eq:vad_fscore}
\end{equation}

\noindent Here, $C_i^{G\rightarrow P}$ measures the coverage of the reference emotion set by the predictions, whereas $C_i^{P\rightarrow G}$ measures the support of the predicted set from the references. The $\beta$-controlled factor explicitly penalizes set-size imbalances. The final score $F_i^{\mathrm{VAD}}$ is their penalty-adjusted harmonic mean.

\subsubsection{Assessment.}
This task evaluates continuous VAD dimension estimation for both expressed and evoked emotions, with predictive fidelity quantified by the Spearman (SRCC) and Pearson (PLCC) correlation coefficients.

\begin{table*}[]
\centering
\setlength{\tabcolsep}{5pt}        
\resizebox{\textwidth}{!}{%
\begin{tabular}{lcccccccccc}
\hline
\multicolumn{1}{c|}{\textbf{Categories}}                       & \multicolumn{10}{c}{\textbf{Evoked}}                                                                                                                                                                                                                                                                                                                                                                                                                                                                                                             \\ \hline
\multicolumn{1}{c|}{\textbf{Tasks}}                            & \multicolumn{3}{c|}{\textbf{Single Video Perception}}                                                                                   & \multicolumn{3}{c|}{\textbf{Video Pair Perception}}                                                                                              & \multicolumn{1}{c|}{}                                                                                          & \multicolumn{3}{c}{\textbf{OV Emotion Recognition}}                                                                               \\ \cline{1-7} \cline{9-11} 
\multicolumn{1}{c|}{\textbf{Models}}                           & \textit{Yes-or-No↑}                  & \textit{What-How↑}                   & \multicolumn{1}{c|}{\textit{Overall↑}}                    & \textit{Yes-or-No↑}                     & \textit{What-How↑}                      & \multicolumn{1}{c|}{\textit{Overall↑}}                       & \multicolumn{1}{c|}{\multirow{-2}{*}{\textit{\begin{tabular}[c]{@{}c@{}}Perception \\ Overall↑\end{tabular}}}} & \textit{$F^{\mathrm{EW}}$↑}                               & \textit{$F^{\mathrm{VAD}}$↑}                              & \textit{Overall↑}                                             \\ \hline

\multicolumn{1}{c|}{Random guess w/o recognition} & 50.00\%             & 30.33\%            & \multicolumn{1}{c|}{40.17\%}           & 50.00\%             & 30.04\%            & \multicolumn{1}{c|}{40.02\%}           & \multicolumn{1}{c|}{40.10\%}                                                                                  & NA                        & NA              & NA         \\ \hline
\multicolumn{10}{l}{{\textit{Open-source audio-only MLLMs}}}                                                                                                                                                                                                                                                                                                                \\ \hdashline
\multicolumn{1}{l|}{Qwen-Audio~\cite{qwen-audio}}  & 49.80\%             & 39.48\%            & \multicolumn{1}{c|}{44.64\%}           & 49.92\%             & 27.61\%            & \multicolumn{1}{c|}{38.76\%}           & \multicolumn{1}{c|}{41.51\%}                                                                                  & 46.73\%                   & 34.81\%        &    40.77\%      \\
\multicolumn{1}{l|}{SALMONN~\cite{salmonn}}         & 49.71\%             & 36.20\%            & \multicolumn{1}{c|}{42.96\%}           & 61.87\%             & 25.51\%            & \multicolumn{1}{c|}{43.69\%}           & \multicolumn{1}{c|}{43.32\%}                                                                                  & 35.77\%                   & 23.97\%                  & 29.87\%
 \\ \hdashline
\multicolumn{10}{l}{{\textit{Open-source video-only MLLMs}}}                                                                                                                                                                                                                                                                                                                \\ \hdashline
\multicolumn{1}{l|}{LLaVA-NeXT-Video-7B~\cite{llava-next}}  & 49.39\%             & 42.92\%            & \multicolumn{1}{c|}{46.15\%}           & 43.43\%             & 31.57\%            & \multicolumn{1}{c|}{37.50\%}           & \multicolumn{1}{c|}{41.88\%}                                                                                  &             54.84\%              &       43.34\%        & 49.09\%           \\

\multicolumn{1}{l|}{VideoChat2~\cite{videochat2}}                & 52.50\%             & 30.88\%            & \multicolumn{1}{c|}{41.69\%}           & 50.42\%             & 28.79\%            & \multicolumn{1}{c|}{39.60\%}           & \multicolumn{1}{c|}{40.66\%}                                                                                  & 57.17\%                   & 44.66\%      & 50.92\%            \\
\multicolumn{1}{l|}{Janus-Pro-7B~\cite{janus}}      & 55.86\%             & 49.39\%            & \multicolumn{1}{c|}{52.62\%}           & 48.48\%             & 33.25\%            & \multicolumn{1}{c|}{40.87\%}           & \multicolumn{1}{c|}{46.82\%}                                                                                  &            56.86\%               &           44.18\%       & 50.52\%        \\

\multicolumn{1}{l|}{InternVL3-8B~\cite{internvl3}}     & 64.05\%             & 60.85\%            & \multicolumn{1}{c|}{62.45\%}           & 61.20\%             & \underline{45.54\%}            & \multicolumn{1}{c|}{53.37\%}           & \multicolumn{1}{c|}{57.97\%}                                                                                  &                \underline{64.02\%}           &               50.54\%    & \underline{57.28\%}       \\
\multicolumn{1}{l|}{Qwen3.5-VL-9B~\cite{qwen3.5}}     & 63.23\%             & 62.08\%            & \multicolumn{1}{c|}{62.65\%}           & 60.77\%             & \textbf{46.21\%}            & \multicolumn{1}{c|}{53.49\%}           & \multicolumn{1}{c|}{58.14\%}                                                                                  &           62.87\%                &         \underline{50.97\%}    & 56.92\%             \\
\multicolumn{1}{l|}{AffectGPT~\cite{AffectGPT}}               & 46.27\%             & 39.07\%            & \multicolumn{1}{c|}{42.67\%}           & 35.52\%             & 29.04\%            & \multicolumn{1}{c|}{32.28\%}           & \multicolumn{1}{c|}{37.55\%}                                                                                  & 36.23\%                   & 27.09\%       & 31.66\%           \\
\multicolumn{1}{l|}{Emotion-Qwen~\cite{emotion-qwen}}                 & 63.72\%             & 60.61\%            & \multicolumn{1}{c|}{62.16\%}           & \textbf{64.73\%}             & 42.68\%            & \multicolumn{1}{c|}{\underline{53.70\%}}           & \multicolumn{1}{c|}{57.99\%}                                                                                  & 55.81\%                   & 44.43\%      & 50.12\%            \\
\multicolumn{1}{l|}{EEmo-Logic~\cite{eemologic}}                   & 65.93\%             & 63.55\%            & \multicolumn{1}{c|}{64.72\%}           & 62.63\%             & 41.25\%            & \multicolumn{1}{c|}{51.94\%}           & \multicolumn{1}{c|}{\underline{58.42\%}}                                                                                  & 56.70\%                   & 44.85\%      & 50.78\%            \\ \hdashline
\multicolumn{10}{l}{{\textit{Open-source Omni-modal MLLMs}}}                                                                                                                                                                                                                                                                                                                      \\ \hdashline
\multicolumn{1}{l|}{R1-Omni-0.5B~\cite{R1-Omni}}             & 57.08\%             & 56.27\%            & \multicolumn{1}{c|}{56.67\%}           & 55.81\%             & 41.08\%            & \multicolumn{1}{c|}{48.44\%}           & \multicolumn{1}{c|}{52.62\%}                                                                                  &           25.65\%                &        15.42\%       & 20.54\%           \\
\multicolumn{1}{l|}{InteractiveOmni-8B~\cite{interactiveomni}}           & 58.59\%             & 62.41\%            & \multicolumn{1}{c|}{60.65\%}           & 50.84\%             & 34.76\%            & \multicolumn{1}{c|}{42.80\%}           & \multicolumn{1}{c|}{51.85\%}                                                                                  & 60.18\%                   & 45.53\%      & 52.86\%            \\
\multicolumn{1}{l|}{Qwen2.5-Omni-7B~\cite{Qwen2.5-Omni}}              & 64.78\%             & \underline{66.01\%}            & \multicolumn{1}{c|}{65.40\%}           & 48.23\%             & 42.76\%            & \multicolumn{1}{c|}{45.50\%}           & \multicolumn{1}{c|}{55.59\%}                                                                                  & 60.51\%                   & 49.36\%       & 54.94\%           \\
\multicolumn{1}{l|}{Qwen3-Omni-30B-A3B~\cite{qwen3-omni}}           & \underline{67.40\%}             & 62.88\%            & \multicolumn{1}{c|}{\underline{65.64\%}}           & \underline{63.80\%}             & 44.95\%            & \multicolumn{1}{c|}{\textbf{54.38\%}}           & \multicolumn{1}{c|}{\textbf{60.09\%}}                                                                                  & 60.51\%                   & 48.68\%      & 54.60\%            \\ \hdashline
\multicolumn{10}{l}{{\textit{Proprietary video-only MLLMs}}}                                                                                                                                                                                                                                                                                                                 \\ \hdashline
\multicolumn{1}{l|}{GPT-5.4~\cite{gpt5.4}}                      & \textbf{67.73\%}             & \textbf{66.18\%}            & \multicolumn{1}{c|}{\textbf{66.95\%}}           & 57.74\%             & 41.16\%            & \multicolumn{1}{c|}{49.45\%}           & \multicolumn{1}{c|}{58.32\%}                                                                                  & \textbf{67.65\%}                   & \textbf{55.38\%}        & \textbf{61.52\%}          \\
\multicolumn{1}{l|}{Claude-4.6-Sonnet~\cite{claude_sonnet_46}}            & 62.57\%             & 64.21\%            & \multicolumn{1}{c|}{63.39\%}           & 52.95\%             & 38.55\%            & \multicolumn{1}{c|}{45.75\%}           & \multicolumn{1}{c|}{54.69\%}                                                                                  & 60.21\%                   & 48.76\%      & 54.49\%            \\ \hline
\end{tabular}%
}
\caption{Performance comparisons on the Perception and Recognition tasks of the evoked-emotion subset of E$^3$mo-Bench. The best performance is \textbf{bolded} 
and the second performance is \underline{underlined}. NA denotes not applicable.
}
\label{tab: evoked}
\vspace{-0.5cm}
\end{table*}

\section{Bayesian Pairwise Alignment Framework}

We introduce BPA as a scalable human-annotation paradigm that replaces costly MOS-based ratings with efficient pairwise judgments. It recovers continuous scores through informative comparison scheduling and Bayesian inference. We further transfer this paradigm to automated scoring, where an MLLM committee serves as a virtual annotator within E$^3$mo-Score.

\subsection{Algorithm Overview}

The proposed algorithm integrates $3$ core modules: online pair scheduling, periodic Bayesian optimization, and uncertainty estimation, as outlined below. Comprehensive theoretical derivations are deferred to the \textit{Supp}.

\subsubsection{Online Pair Scheduling.}

BPA characterizes each sample via a score $\mu$ and an uncertainty $\sigma$. For target samples, the initial score $\mu^{(0)}$ is derived from the bootstrap preliminary VAD estimates, accompanied by a large initial uncertainty $\sigma^{(0)}$. Conversely, Golden Anchors serve as fixed references with static $\mu$ and near-zero $\sigma$. To simultaneously optimize information gain, scale calibration, and graph connectivity, we implement a tripartite sampling policy inspired by~\cite{hybridmst}: (1) Exploration ($50\%$), pairing high-uncertainty samples with similar $\mu$ to maximize information gain via TrueSkill~\cite{trueskill}; (2) Anchor Calibration ($30\%$), matching samples against proximal anchors to correct scale drift~\cite{anchor}; and (3) Bridge Sampling ($20\%$), connecting samples across predefined levels to prevent graph fragmentation. 
After each comparison, the original online rule provisionally updates $\mu$ and $\sigma$ solely to guide short-term scheduling.

\subsubsection{Periodic Bayesian Optimization.}

Upon accumulating a predefined threshold of new comparisons, BPA jointly optimizes the latent scores across all samples. Letting $\mu_i$ denote the latent score of sample $i$ for a given affective dimension, the score difference for pair $(i,j)$ is $\Delta_{ij}=\mu_i-\mu_j$. To explicitly account for perceptually indistinguishable pairs, we adopt the Rao–Kupper formulation~\cite{rao-kupper} with a tie margin $\delta>0$. The respective probabilities for $i$ winning, $j$ winning, or a tie are formulated as follows:

\vspace{-0.2cm}
\begin{equation}
\begin{gathered}
p_{ij}^{+}
=
\Lambda\!\left(\Delta_{ij}-\delta\right),
\qquad
p_{ij}^{-}
=
\Lambda\!\left(-\Delta_{ij}-\delta\right),
\\
p_{ij}^{0}
=
\Lambda\!\left(\Delta_{ij}+\delta\right)
-
\Lambda\!\left(\Delta_{ij}-\delta\right),
\end{gathered}
\label{eq:rao_kupper_probabilities}
\end{equation}

\noindent where $\Lambda(z)=\left(1+\exp(-z)\right)^{-1}$ denotes the logistic sigmoid function. Inspired by Crowd-BT~\cite{crowdBT}, annotator reliability is modeled by $\eta_k\in[0,1]$ and updated after each optimization round (see \textit{Supp}). The annotator either follows the true latent Rao–Kupper distribution with probability $\eta_k$, or otherwise outputs a random guess. Thus, for an observed outcome $y\in\{+,0,-\}$, the likelihood is formulated as:

\vspace{-0.2cm}
\begin{equation}
    q_{ijk}^{y}=P(y\mid \mu_{i},\mu_{j},\eta_{k},\delta)=\eta_{k}p_{ij}^{y}+\frac{1-\eta_{k}}{3}.
\end{equation}

\noindent This mixture formulation effectively downweights unreliable judgments while preserving explicit win, loss, and tie observations. Given $M$ pairwise comparison records, the base negative log-likelihood is defined as follows:

\vspace{-0.2cm}
\begin{equation}
    \mathcal{L}_{\mathrm{data}}=-\sum_{m=1}^M\log q_{i_mj_mk_m}^{y_m}.
\end{equation}

\noindent Let $\mathcal{G}$ denote the Golden Anchor Set. The scores of these anchors are fixed to their direct human ratings throughout optimization. For each candidate sample $i$, an adaptive quadratic prior is attached to the initial reference score $\mu_i^{(0)}$:

\begin{equation}
    \mathcal{L}_{\mathrm{prior}} = \sum_{i\notin\mathcal{G}}
\frac{\lambda_0}{1+\gamma N_i}
\left(
\mu_i-\mu_i^{(0)}
\right)^2,
\end{equation}




\noindent where $N_i$ denotes the number of comparisons involving sample $i$, $\lambda_0$ controls the base prior strength, and $\gamma$ governs its decay with increasing evidence. The prior is therefore stronger for sparsely compared samples and weakens as more comparisons are collected. The complete MAP objective is $\mathcal{L}_{\mathrm{total}}=\mathcal{L}_{\mathrm{data}}+\mathcal{L}_{\mathrm{prior}}$. Optimization is rerun periodically after a fixed number of new comparisons to jointly update candidate scores, annotator reliabilities, and the tie margin. Candidate scores are constrained to $[1,9]$, while Golden Anchor scores remain fixed throughout optimization.

\begin{table*}[]
\centering
\setlength{\tabcolsep}{5pt}        
\resizebox{\textwidth}{!}{%
\begin{tabular}{lcccccccccc}
\hline
\multicolumn{1}{c|}{\textbf{Categories}}                       & \multicolumn{10}{c}{\textbf{Expressed}}                                                                                                                                                                                                                                                                                                                                                                                                                                                                                                             \\ \hline
\multicolumn{1}{c|}{\textbf{Tasks}}                            & \multicolumn{3}{c|}{\textbf{Single Video Perception}}                                                                                   & \multicolumn{3}{c|}{\textbf{Video Pair Perception}}                                                                                              & \multicolumn{1}{c|}{}                                                                                          & \multicolumn{3}{c}{\textbf{OV Emotion Recognition}}                                                                               \\ \cline{1-7} \cline{9-11} 
\multicolumn{1}{c|}{\textbf{Models}}                           & \textit{Yes-or-No↑}                  & \textit{What-How↑}                   & \multicolumn{1}{c|}{\textit{Overall↑}}                    & \textit{Yes-or-No↑}                     & \textit{What-How↑}                      & \multicolumn{1}{c|}{\textit{Overall↑}}                       & \multicolumn{1}{c|}{\multirow{-2}{*}{\textit{\begin{tabular}[c]{@{}c@{}}Perception \\ Overall↑\end{tabular}}}} & \textit{$F^{\mathrm{EW}}$↑}                               & \textit{$F^{\mathrm{VAD}}$↑}                              & \textit{Overall↑}                                             \\ \hline
\multicolumn{1}{c|}{Random guess w/o recognition} & 50.00\%             & 30.46\%            & \multicolumn{1}{c|}{40.23\%}           & 50.00\%             & 30.05\%            & \multicolumn{1}{c|}{40.02\%}           & \multicolumn{1}{c|}{40.13\%}                                                                                  & NA                        & NA             &    NA      \\ \hline
\multicolumn{10}{l}{{\textit{Open-source audio-only MLLMs}}}                                                                                                                                                                                                                                                                                                                \\ \hdashline
\multicolumn{1}{l|}{Qwen-Audio~\cite{qwen-audio}}       & 47.28\%             & 37.17\%            & \multicolumn{1}{c|}{42.22\%}           & 30.30\%             & 12.53\%            & \multicolumn{1}{c|}{21.42\%}           & \multicolumn{1}{c|}{32.29\%}                                                                                  & 38.45\%                   & 30.57\%        & 34.54\%          \\
\multicolumn{1}{l|}{SALMONN~\cite{salmonn}}        & 51.35\%             & 35.03\%            & \multicolumn{1}{c|}{43.19\%}           & 60.40\%             & 27.55\%            & \multicolumn{1}{c|}{43.98\%}           & \multicolumn{1}{c|}{43.57\%}                                                                                  & 31.41\%                   & 26.94\%               & 29.18\%  
 \\ \hdashline
\multicolumn{10}{l}{{\textit{Open-source video-only MLLMs}}}                                                                                                                                                                                                                                                                                                                \\ \hdashline
\multicolumn{1}{l|}{LLaVA-NeXT-Video-7B~\cite{llava-next}}     & 58.12\%             & 50.81\%            & \multicolumn{1}{c|}{54.46\%}           & 44.23\%             & 31.42\%            & \multicolumn{1}{c|}{37.83\%}           & \multicolumn{1}{c|}{46.52\%}                                                                                  &           51.19\%                &      43.98\%       & 47.59\%             \\
\multicolumn{1}{l|}{VideoChat2~\cite{videochat2}}           & 51.66\%             & 32.87\%            & \multicolumn{1}{c|}{42.26\%}           & 49.20\%             & 29.74\%            & \multicolumn{1}{c|}{39.47\%}           & \multicolumn{1}{c|}{40.93\%}                                                                                  & 41.99\%                   & 35.25\%       & 38.62\%           \\
\multicolumn{1}{l|}{Janus-Pro-7B~\cite{janus}}       & 60.51\%             & 54.43\%            & \multicolumn{1}{c|}{57.47\%}           & 50.13\%             & 34.54\%            & \multicolumn{1}{c|}{42.33\%}           & \multicolumn{1}{c|}{50.24\%}                                                                                  &            53.79\%               &            46.21\%    & 50.00\%          \\
\multicolumn{1}{l|}{InternVL3-8B~\cite{internvl3}}       & 69.21\%             & 64.51\%            & \multicolumn{1}{c|}{66.86\%}           & 63.35\%             & \textbf{46.50\%}            & \multicolumn{1}{c|}{\textbf{54.93\%}}           & \multicolumn{1}{c|}{61.16\%}                                                                                  &             \underline{55.57\%}              &              48.41\%     & \underline{51.99\%}       \\
\multicolumn{1}{l|}{Qwen3.5-VL-9B~\cite{qwen3.5}}         & \textbf{70.82\%}             & \underline{68.13\%}            & \multicolumn{1}{c|}{\underline{69.48\%}}           & 62.85\%             & \textbf{46.50\%}            & \multicolumn{1}{c|}{54.68\%}           & \multicolumn{1}{c|}{\textbf{62.41\%}}                                                                                  &        55.19\%                   &             \underline{48.43\%}      & 51.81\%       \\
\multicolumn{1}{l|}{AffectGPT~\cite{AffectGPT}}        & 47.11\%             & 42.19\%            & \multicolumn{1}{c|}{44.65\%}           & 40.52\%             & 29.15\%            & \multicolumn{1}{c|}{34.84\%}           & \multicolumn{1}{c|}{39.97\%}                                                                                  & 36.94\%                   & 29.41\%      & 33.18\%            \\
\multicolumn{1}{l|}{Emotion-Qwen~\cite{emotion-qwen}}                 & 67.13\%             & 60.20\%            & \multicolumn{1}{c|}{63.66\%}           & \underline{66.13\%}             & 42.21\%            & \multicolumn{1}{c|}{54.13\%}           & \multicolumn{1}{c|}{59.11\%}                                                                                  & 45.49\%                   & 37.21\%      & 41.35\%            \\
\multicolumn{1}{l|}{EEmo-Logic~\cite{eemologic}}                   & 64.05\%             & 59.74\%            & \multicolumn{1}{c|}{61.89\%}           & 60.66\%             & 43.64\%            & \multicolumn{1}{c|}{52.15\%}           & \multicolumn{1}{c|}{57.24\%}                                                                                  & 48.12\%                   & 42.61\%      & 45.37\%            \\ \hdashline
\multicolumn{10}{l}{{\textit{Open-source Omni-modal MLLMs}}}                                                                                                                                                                                                                                                                                                                      \\ \hdashline
\multicolumn{1}{l|}{R1-Omni-0.5B~\cite{R1-Omni}}               & 62.12\%             & 56.20\%            & \multicolumn{1}{c|}{59.16\%}           & 58.55\%             & 43.64\%            & \multicolumn{1}{c|}{51.10\%}           & \multicolumn{1}{c|}{55.31\%}                                                                                  &             32.50\%              &         22.96\%      & 27.73\%           \\
\multicolumn{1}{l|}{InteractiveOmni-8B~\cite{interactiveomni}}           & 65.28\%             & 65.82\%            & \multicolumn{1}{c|}{65.55\%}           & 52.49\%             & 37.41\%            & \multicolumn{1}{c|}{44.95\%}           & \multicolumn{1}{c|}{55.71\%}                                                                                  & 55.17\%                   & 46.44\%       & 50.81\%           \\
\multicolumn{1}{l|}{Qwen2.5-Omni-7B~\cite{Qwen2.5-Omni}}              & 64.43\%             & 61.05\%            & \multicolumn{1}{c|}{62.74\%}           & 51.14\%             & 41.87\%            & \multicolumn{1}{c|}{46.50\%}           & \multicolumn{1}{c|}{54.99\%}                                                                                  & 52.50\%                   & 44.47\%        & 48.49\%          \\
\multicolumn{1}{l|}{Qwen3-Omni-30B-A3B~\cite{qwen3-omni}}           & 69.52\%             & 65.51\%            & \multicolumn{1}{c|}{67.51\%}           & \textbf{66.22\%}             & 43.30\%            & \multicolumn{1}{c|}{\underline{54.76\%}}           & \multicolumn{1}{c|}{61.42\%}                                                                                  & 53.94\%                   & 45.88\%     & 49.91\%             \\ \hdashline
\multicolumn{10}{l}{{\textit{Proprietary video-only MLLMs}}}                                                                                                                                                                                                                                                                                                                 \\ \hdashline
\multicolumn{1}{l|}{GPT-5.4~\cite{gpt5.4}}                      & \underline{70.67\%}             & \textbf{70.21\%}            & \multicolumn{1}{c|}{\textbf{70.44\%}}           & 53.31\%             & \underline{44.90\%}            & \multicolumn{1}{c|}{52.11\%}           & \multicolumn{1}{c|}{\underline{61.69\%}}                                                                                  & \textbf{57.73\%}                   & \textbf{48.89\%}      & \textbf{53.31\%}            \\
\multicolumn{1}{l|}{Claude-4.6-Sonnet~\cite{claude_sonnet_46}}            & 66.97\%             & 67.67\%            & \multicolumn{1}{c|}{67.32\%}           & 54.51\%             & 40.35\%            & \multicolumn{1}{c|}{47.43\%}           & \multicolumn{1}{c|}{57.82\%}  & 52.04\%                   & 43.78\%       & 47.91\%           \\ \hline
\end{tabular}%
}
\caption{Performance comparisons on the Perception and Recognition tasks of the expressed-emotion subset of E$^3$mo-Bench.}
\label{tab: expressed}
\vspace{-0.5cm}
\end{table*}

\subsubsection{Uncertainty Estimation.}

To estimate candidate uncertainty $\sigma$ after each global MAP update, we leverage a diagonal empirical-Fisher approximation~\cite{fisher1,fisher2}. By explicitly summing the information from the observed pairwise likelihoods and the adaptive prior precision, we derive:

\vspace{-0.2cm}
\begin{equation}
I_i=\sum_{m\in\mathcal{D}_i}\left(\frac{\partial\log q_{i_mj_mk_m}^{y_m}}{\partial\mu_i}\right)^2+\frac{2\lambda_0}{1+\gamma N_i},\widehat{\sigma}_i=I_i^{-1/2}.
\end{equation}

\noindent To prevent abrupt fluctuations across consecutive offline updates, the raw Fisher estimate is temporally smoothed with the preceding uncertainty and clipped to a predefined range:

\vspace{-0.2cm}
\begin{equation}
\sigma_i^{\mathrm{new}}=\mathrm{clip}\left(\rho\widehat{\sigma}_i+(1-\rho)\sigma_i^{\mathrm{old}},\sigma^{\min},\sigma^{(0)} \right),
\end{equation}

\noindent where $\rho\in[0,1]$ dictates the update rate and $\sigma^{\min}$ sets the uncertainty floor. The clipped $\sigma_i^{\mathrm{new}}$ functions as a diagonal Laplace approximation of the local posterior uncertainty. Iterative annotation halts once the global mean uncertainty of candidate samples drops below the stopping threshold $\sigma_\tau$.

\subsection{Human Pairwise Annotation}

We deploy the proposed BPA framework to collect pairwise VAD judgments for the Broad-Coverage Set. To mitigate cognitive load and cross-dimensional interference, we divide the expressed and evoked samples into $6$ independent subtasks, restricting participants to one dimension each. This pipeline yields $35{,}639$ valid annotations. To verify annotation fidelity, we relax the Golden Anchor constraints during optimization; the re-estimated anchor scores align strongly with original ratings. Especially, BPA incurs approximately $60\%$ of the cost of conventional MOS protocols, demonstrating its efficacy for reliable and scalable VAD annotation. Complete validation procedures and results are provided in \textit{Supp}.

\subsection{The E$^3$mo-Score Agent}

Extending the BPA paradigm, we present E$^3$mo-Score (Fig.~\ref{fig: BPA agent}), a model-driven scoring agent. Since uni-dimensional comparisons bypass the steep reasoning overhead of direct VAD regression, we employ a model committee to harness complementary model strengths. The ensemble consists of $3$ general-purpose multimodal models, \textit{Qwen3-Omni}~\cite{qwen3-omni}, \textit{Qwen2.5-Omni}~\cite{Qwen2.5-Omni}, and \textit{InteractiveOmni}~\cite{interactiveomni}, together with $2$ emotion-oriented models, \textit{EEmo-Logic}~\cite{eemologic} and \textit{Emotion-Qwen}~\cite{emotion-qwen}. Evaluators leverage their native modalities, supplemented by text-summarized auditory cues when native audio is lacking. To ensure reliable calibration, a subset of the Golden Anchors is used to estimate each model’s confidence weight for every subtask. Following weighted voting, final judgments are fed into the identical BPA optimization pipeline used for human annotators, converging under the same termination criteria to yield final VAD predictions for the entire batch.

\section{Experiments}

\subsection{Experimental Setup}

In this section, we evaluate $14$ open-source and $2$ proprietary MLLMs on E$^3$mo-Bench. Based on their supported input modalities, the models are grouped into audio-only, video-only, and omni-modal settings, covering both general-purpose and emotion-oriented MLLMs. We further evaluate the proposed E$^3$mo-Score Agent on the Golden Anchor Set. Detailed descriptions of the evaluated MLLMs, BPA hyperparameter settings, and additional analyses are provided in \textit{Supp}.

\subsection{Evaluation on Perception Task}

\subsubsection{Evoked Emotion Perception.}

As presented in Tab.~\ref{tab: evoked} and Fig.~\ref{fig:radar}(a), Qwen3-Omni achieves the highest overall performance ($60.09\%$), underscoring the necessity of joint audio-visual analysis for evoked emotion modeling. EEmo-Logic ranks second ($58.42\%$), validating the strong transferability of task-specific knowledge acquired from image-evoked emotion datasets. At a granular level, GPT-5.4 excels in single-video scenarios ($66.95\%$) due to its superior visual comprehension. For complex video pairs, however, Qwen3-Omni leads with only $54.38\%$, indicating significant room for improvement in robust evoked emotion understanding.

\subsubsection{Expressed Emotion Perception.}

\begin{figure}
    \centering
    \includegraphics[width=\linewidth]{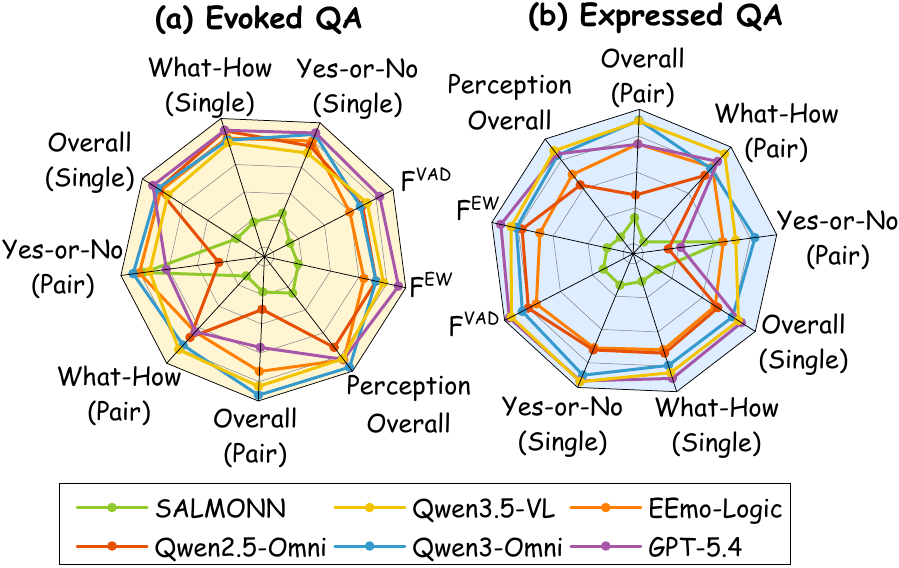}
    \caption{Performance on Perception and Recognition tasks of Evoked and Expressed Emotions.}
    \label{fig:radar}
    \vspace{-0.4cm}
\end{figure}

As detailed in Tab.~\ref{tab: expressed} and Fig.~\ref{fig:radar}(b), the vision-centric models Qwen3.5-VL and GPT-5.4 achieve top overall performance at $62.41\%$ and $61.69\%$, respectively. This highlights visual cues, particularly facial expressions and body movements, as the primary evidence for inferring expressed emotions. For single-video tasks, GPT-5.4 maintains its dominance ($70.44\%$), whereas InternVL3 leads in video-pair analysis ($54.93\%$). Notably, the strong performance of recent mid-scale open-source models highlights that their training corpora are heavily optimized for the understanding of human-centric affective expressions.

\subsubsection{Overall Analysis.}

Synthesizing these findings yields $3$ key insights: 1) Video-pair analysis is significantly more challenging than single-video evaluation. With peak performance exceeding random chance by a mere $14.63\%$, current MLLMs clearly struggle with fine-grained affective modeling and multiplexed cue processing. 2) Models perceive expressed emotions slightly better than evoked emotions; however, as both are foundational to EI, they require synergistic advancement. 3) Although audio-only models lag and video-only results cement vision as the primary affective modality, omni-modal architectures demonstrate the highest ceiling for holistic emotion understanding, charting a definitive course for future research. Additional empirical details are provided in \textit{Supp.}

\subsection{Evaluation on Recognition Task}

Tabs.~\ref{tab: evoked} and ~\ref{tab: expressed} reveal $3$ key insights regarding OV emotion recognition: (1) Current MLLMs exhibit stronger fine-grained recognition for evoked emotions, indicating that the multiplexed cues inherent in expressed emotions (e.g., facial micro-expressions) impose steeper modeling challenges. (2) While $F^{(EW)}$ and $F^{(VAD)}$ correlate strongly, $F^{(VAD)}$ uniquely quantifies nuanced differences in error severity. For instance, despite achieving identical $F^{(EW)}$ scores on evoked emotions, Qwen2.5-Omni outperforms Qwen3-Omni in $F^{(VAD)}$, proving that its misclassifications incur lower semantic penalties and reflect deeper reasoning. (3) The state-of-the-art results of GPT-5.4 ($61.52\%$ and $53.31\%$) and InternVL3 ($57.28\%$ and $51.99\%$) across both perspectives reaffirm vision as the dominant modality for affective inference. Ultimately, these performance ceilings expose a persistent gap in true fine-grained comprehension, guiding future multimodal research.

\subsection{Evaluation on Assessment Task}

Because most baseline models, except EEmo-Logic, cannot directly produce calibrated VAD scores, we adopt the probability-based scoring protocol of Q-Bench~\cite{qbench}, with details provided in \textit{Supp}. To validate E$^3$mo-Score, we retain a small subset of the Golden Anchor Set as fixed anchors and infer the remaining scores through BPA. Tab.~\ref{tab: assessment} compares their correlations with human ratings against those achieved by the $5$ committee models. E$^3$mo-Score outperforms every individual model across all metrics except expressed valence, demonstrating the robustness of BPA-based consensus. Among the individual models, Qwen3-Omni achieves the strongest overall performance, suggesting that large-scale omni-modal integration better captures fine-grained affective intensity. Nevertheless, dominance remains challenging for all methods. Evaluation on the full E$^3$mo-Bench yields consistent findings, with detailed results, analyses, and ablation studies provided in \textit{Supp}.

\begin{table}[]
\centering
\setlength{\tabcolsep}{2pt}        
\resizebox{\linewidth}{!}{
\begin{tabular}{l|ccc|ccc}
\hline
\multicolumn{1}{c|}{\textbf{Categories}} & \multicolumn{3}{c|}{\textbf{Evoked}}                         & \multicolumn{3}{c}{\textbf{Expressed}}                       \\ \hline
\multicolumn{1}{c|}{\textbf{Dimensions}} & \textbf{V.}   & \textbf{A.}   & \textbf{D.} & \textbf{V.}   & \textbf{A.}   & \textbf{D.} \\ \hline
InteractiveOmni                       & 0.52/0.51          & 0.45/0.45          & 0.03/0.09          & 0.58/0.60          & 0.48/0.49          & 0.35/0.38          \\
Emotion-Qwen                             & 0.48/0.47          & 0.40/0.39          & 0.14/0.16          & 0.57/0.58          & 0.48/0.48          & 0.40/0.42          \\
EEmo-Logic                               & 0.55/0.55          & 0.47/0.48          & 0.45/0.44          & 0.58/0.60          & 0.30/0.32          & 0.43/0.44          \\
Qwen2.5-Omni                          & 0.68/0.66          & 0.64/0.66          & 0.28/0.25          & 0.54/0.57          & 0.54/0.57          & 0.26/0.27          \\
Qwen3-Omni                       & 0.75/0.74          & 0.66/0.68          & 0.43/0.39          & \textbf{0.77/0.75} & 0.65/0.67          & 0.53/0.52          \\
\textbf{E$^3$mo-Score (Ours)}                               & \textbf{0.76/0.77} & \textbf{0.71/0.70} & \textbf{0.62/0.60} & 0.73/0.74          & \textbf{0.73/0.75} & \textbf{0.61/0.62} \\ \hline
\end{tabular}%
}
\caption{Performance comparison on the E$^3$mo-Bench Assessment task. Metrics: \textit{SRCC}↑ / \textit{PLCC}↑.}
\label{tab: assessment}
\end{table}

\begin{table}[]
\setlength{\tabcolsep}{4pt}        
\resizebox{\linewidth}{!}{
\begin{tabular}{c|c|ccc|cc}
\hline
               &                  & \multicolumn{3}{c|}{\textbf{Pair Dispatch}}                                                 & \multicolumn{2}{c}{\textbf{Assessment}} \\ \cline{3-7}  \multirow{-2}{*}{No.} &
\multirow{-2}{*}{\textbf{Optimization}} & TrueSkill & \begin{tabular}[c]{@{}c@{}}Anchor \\ Calibration\end{tabular} & Bridge & \textit{SRCC↑}           & \textit{PLCC↑}          \\ \hline
(1)                   & TrueSkill                      & \checkmark         &                    &        &      0.74          &      0.73         \\
(2)                   &Bayesian                            & \checkmark         &                    &        &        0.73        &         0.72      \\
(3)                   &Bayesian                            & \checkmark         & \checkmark                  &        &        0.68        &      0.67         \\
(4)                   &Bayesian                            & \checkmark         &                    & \checkmark      &          0.75      &      0.74         \\
\rowcolor[HTML]{D9D9D9} 
(5)                   &Bayesian                            & \checkmark         & \checkmark                  & \checkmark      & \textbf{0.76}           & \textbf{0.77}          \\ \hline
\end{tabular}
}
\caption{Ablation studies of the optimization and pair-dispatch strategies in E$^3$mo-Score on the evoked-valence dimension.}
\label{tab: ablation}
\vspace{-0.4cm}
\end{table}

\subsection{Ablation Study}

Tab.~\ref{tab: ablation} ablates the BPA optimization and pair-dispatch strategies within E$^3$mo-Score, using a unified stopping criterion whereby optimization terminates once the mean uncertainty across candidate samples falls below $\sigma_\tau$. Comparing rows $(1)$ and $(2)$ shows that standard TrueSkill slightly outperforms its Bayesian-refined counterpart, suggesting that purely TrueSkill dispatch provides insufficient bridge comparisons to fully exploit BPA. However, standard TrueSkill's inability to assess the reliability of individual judgments limits its suitability as the core inference algorithm for manual annotation. Rows $(2)$ through $(5)$ equalize the proportions of various dispatch strategies, confirming that our default configuration attains optimal results. Specifically, TrueSkill’s uncertainty-aware dispatch drives the primary performance gain, while anchor calibration and bridge sampling jointly provide additional improvements by stabilizing the global score structure. Additional experiments and analyses are detailed in \textit{Supp}. 

\section{Conclusion}

In this paper, we introduce \textbf{E$^3$mo-Bench}, a comprehensive benchmark for evaluating MLLMs on both evoked and expressed emotion understanding. The benchmark is structured around $3$ complementary tasks: emotion perception, OV emotion recognition, and VAD assessment. To support reliable annotation at scale, we propose \textbf{BPA}, a scalable framework that infers globally consistent dimensional scores from pairwise judgments. Building upon BPA, we further develop the training-free \textbf{E$^3$mo-Score} agent, which aggregates complementary decisions from a five-model committee to produce score predictions. Extensive experiments validate the effectiveness of our frameworks while exposing persistent MLLM limitations, particularly in pairwise perception and fine-grained recognition. Overall, E$^3$mo-Bench and the BPA-based scoring framework provide a systematic testbed for evaluating and advancing multimodal emotional intelligence.

\clearpage
\bibliography{reference}


\end{document}


\maketitle

\section{Overview} \label{sec: overview}

The supplementary material provides detailed descriptions of the data, methodology, experimental setup, and additional results. It is organized as follows. \textbf{Sec.~\ref{sec: dataset comparison}} presents the question-answering~(QA) pairs distribution of E$^3$mo-Bench and compares it with related datasets and benchmarks. \textbf{Sec.~\ref{sec: construction}} details the benchmark construction process, including the prompts used to query multimodal large language models (MLLMs), annotation quality control, and reliability analyses. \textbf{Sec.~\ref{sec: algorithm}} provides further methodological details, background on the underlying algorithms, and the protocols used for MLLM evaluation. \textbf{Sec.~\ref{sec: implementation details}} reports the implementation details, including descriptions of the evaluated models and the hyperparameter settings of Bayesian pairwise alignment (BPA). \textbf{Sec.~\ref{sec: additional results}} presents additional experimental results and analyses, including a category-wise analysis of the dimensions targeted by the QA pairs, complete results for the Assessment task, and further ablation studies of E$^3$mo-Score. \textbf{Sec.~\ref{sec: qualitative results}} provides additional qualitative visualizations, and \textbf{Sec.~\ref{sec: limitations}} discusses the limitations of this work.

\section{Benchmark Statistics and Comparisons} \label{sec: dataset comparison}

\subsection{Benchmark Statistics}

\begin{figure}
    \centering
    \includegraphics[width=\linewidth]{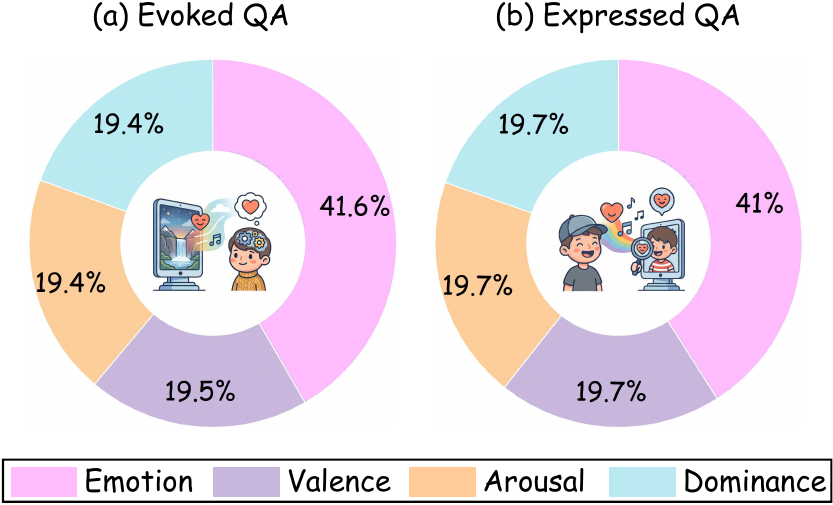}
    \caption{Distribution of question focus dimensions in the Perception task for (a) evoked and (b) expressed emotions.}
    \label{fig:QA distribution}
\end{figure}

Here, we detail the QA composition of E$^3$mo-Bench. For the Perception task, we design $20$ templates for individual videos and $16$ for video pairs, covering $4$ affective dimensions: \textbf{Emotion}, \textbf{Valence}, \textbf{Arousal}, and \textbf{Dominance}. The QA pairs are generated through a rule-based, model-assisted pipeline based on the predefined emotional labels. Each video is paired with one \textit{Yes-or-No} question and one \textit{What-How} question, while a comparable number of pairwise questions are formulated to facilitate comparative analysis. The evoked-emotion subset contains $2{,}442$ single-video QA pairs and $2{,}375$ video-pair QA pairs, whereas the expressed-emotion subset contains $2{,}598$ and $2{,}375$, respectively. The distribution across affective dimensions is shown in Fig.~\ref{fig:QA distribution}. In addition, we construct one open-vocabulary (OV) emotion-reasoning QA for each video to evaluate fine-grained emotion perception and recognition by MLLMs.

\begin{table*}[]
\centering
\setlength{\tabcolsep}{3pt}        
\resizebox{\textwidth}{!}{%
\begin{tabular}{lccccccccc}
\hline
\textbf{Datasets} & \textbf{Modality}      & \textbf{\begin{tabular}[c]{@{}c@{}}Emotion\\ Categories\end{tabular}} & \textbf{\begin{tabular}[c]{@{}c@{}}Emotion\\ Diversity\end{tabular}} & \textbf{\begin{tabular}[c]{@{}c@{}}Emotion\\ Dimension\end{tabular}} & \textbf{\begin{tabular}[c]{@{}c@{}}QA\\ pairs\end{tabular}} & \textbf{\begin{tabular}[c]{@{}c@{}}Pairwise\\ Samples\end{tabular}} & \textbf{\begin{tabular}[c]{@{}c@{}}Reasoning\\ Description\end{tabular}} & \textbf{Scale} & \textbf{Annotation}      \\ \hline
\multicolumn{10}{l}{\textit{Focuses on expressed emotion}}                                                                                                                                                                                                                                                                                                                                                                                                                                                                  \\ \hdashline
MELD~\cite{MELD}         & Video \& Audio \& Text & 7                                                                     & \xmark                                                                    & \xmark                                                                    & \xmark                                                           & \xmark                                                                   & \xmark                                                                        & 13,707         & Human                    \\
CAER~\cite{CAER}          & Video \& Audio \& Text & 7                                                                     & \xmark                                                                    & \xmark                                                                    & \xmark                                                           & \xmark                                                                   & \xmark                                                                        & 13,201         & Human                    \\
DFEW~\cite{DFEW}          & Video \& Audio \& Text & 7                                                                     & \xmark                                                                    & \xmark                                                                    & \xmark                                                           & \xmark                                                                   & \xmark                                                                        & 16,372         & Human                    \\
MAFW~\cite{MAFW}          & Video \& Audio \& Text & 11                                                                    & \checkmark                                                                    & \xmark                                                                    & \xmark                                                           & \xmark                                                                   & \xmark                                                                        & 10,045         & Human                    \\
OV-MER~\cite{OV-MER}        & Video \& Audio \& Text & OV                                                                    & \checkmark                                                                    & \xmark                                                                    & \xmark                                                           & \xmark                                                                   & \checkmark                                                                        & 322            & Model-led+Human-assisted \\
EmoBench-M~\cite{emobench-m}    & Video \& Audio \& Text & 21                                                                    & \xmark                                                                    & \xmark                                                                    & 6,226                                                       & \xmark                                                                   & \xmark                                                                        & 5,642          & Human (Source)           \\
MME-Emotion~\cite{mme-emotion}   & Video \& Audio \& Text & 33                                                                    & \checkmark                                                                    & \xmark                                                                    & 6,500                                                       & \xmark                                                                   & \xmark                                                                        & 6,500          & Human (Source)           \\
MER-Caption~\cite{AffectGPT}   & Video \& Audio \& Text & OV                                                                    & \checkmark                                                                    & \xmark                                                                    & \xmark                                                           & \xmark                                                                   & \checkmark                                                                        & 115,595        & Model-led+Human-assisted \\
EmoReAlM~\cite{AVERE}      & Video \& Audio \& Text & 7                                                                     & \xmark                                                                    & \xmark                                                                    & 4,000                                                       & 41,687                                                              & \checkmark                                                                        & 2,649          & Model-led+Human-assisted \\ \hdashline
\multicolumn{10}{l}{\textit{Focuses on evoked emotion}}                                                                                                                                                                                                                                                                                                                                                                                                                                                                     \\ \hdashline
VideoEmotion~\cite{VideoEmotion}  & Video \& Audio \& Text & 8                                                                     & \xmark                                                                    & \xmark                                                                    & \xmark                                                           & \xmark                                                                   & \xmark                                                                        & 1,099          & Human                    \\
LIRIS-ACCEDE~\cite{LIRIS-ACCEDE} & Video \& Audio \& Text & \xmark                                                                     & \xmark                                                                    & VA                                                                   & \xmark                                                           & \xmark                                                                   & \xmark                                                                        & 9,800          & Human                    \\
MediaEval~\cite{MediaEval}    & Video \& Audio \& Text & \xmark                                                                     & \xmark                                                                    & VA                                                                   & \xmark                                                           & \xmark                                                                   & \xmark                                                                        & 1,201          & Human                    \\
VAD Danmu~\cite{vad-danmu}          & Video \& Audio \& Text & 13                                                                    & \xmark                                                                    & VA                                                                   & \xmark                                                           & \xmark                                                                   & \xmark                                                                        & 19,267         & Human                    \\
EEmo-Bench~\cite{eemo-bench}    & Image \& Text          & 7                                                                     & \checkmark                                                                    & VAD                                                                  & 6,773                                                       & 1,008                                                               & \checkmark                                                                        & 1,960          & Human                    \\ \hline
\multicolumn{10}{l}{\textit{Covers both expressed and evoked emotions}}                                                                                                                                                                                                                                                                                                                                                                                                                                                     \\ \hdashline
AICA-Bench~\cite{AICA-Bench}    & Image \& Text          & 41                                                                    & \xmark                                                                    & \xmark                                                                    & 18,124                                                      & \xmark                                                                   & \checkmark                                                                        & 8,086          & Model-led+Human-assisted \\
\textbf{E$^3$mo-Bench (Ours)} & Video \& Audio \& Text & OV                                                                    & \checkmark                                                                    & VAD                                                                  & 12,314                                                      & 2,375                                                               & \checkmark                                                                        & 2,524          & Human-led+Model-assisted \\ \hline
\end{tabular}
}
\caption{Comparison of E$^3$mo-Bench with other emotion datasets and benchmarks.}
\label{tab: dataset comparison}
\end{table*}

\subsection{Details of Related Emotion Datasets}
\textbf{MELD} \cite{MELD} is a multimodal, multi-party conversational dataset for emotion recognition, derived from dialogues in the television series \textit{Friends}. It comprises approximately $13{,}000$ utterances across $1{,}433$ dialogues, each involving multiple speakers. Every utterance is accompanied by synchronized audio, visual, and textual modalities, and is annotated with $7$ discrete emotion categories—anger, disgust, fear, joy, sadness, surprise, and neutral—as well as three sentiment polarities (positive, negative, and neutral).

\textbf{CAER} \cite{CAER} is a video dataset for context-aware emotion recognition, comprising approximately $13{,}201$ clips collected from $79$ TV shows, totaling around $1.1$ million frames. Each clip is independently annotated by three annotators with $7$ emotion categories (anger, disgust, fear, happy, neutral, sadness, surprise). Unlike datasets providing only cropped facial images, CAER includes full scenes and human body regions, enabling joint modeling of facial expressions and contextual cues.

\textbf{DFEW} \cite{DFEW} is a large-scale video dataset for dynamic facial expression recognition, comprising approximately $16{,}000$ clips extracted from over $1{,}500$ movies. Each clip is independently annotated by $10$ annotators with seven basic emotion categories. The data covers real-world challenges including extreme illumination, occlusions, and pose variations, with substantial annotation quality (Fleiss' kappa~\cite{fleiss1971measuring} of $0.70$). Both single-label annotations and seven-dimensional emotion distributions are provided.

\textbf{MAFW} \cite{MAFW} is a large-scale multimodal compound affective dataset comprising $10{,}045$ video-audio clips collected from diverse sources such as movies, TV series, and online short videos. Each clip is independently annotated by $11$ annotators, covering $11$ single emotion categories and $32$ multi-label compound emotion categories. The dataset also provides bilingual descriptive texts documenting facial action units, body movements, and environmental cues. It is the first dynamic multimodal affective database that offers both compound emotion annotations and emotion-related captions.

\textbf{OV-MER} \cite{OV-MER} is a dataset for OV multimodal emotion recognition, comprising samples selected and re-annotated from MER2023~\cite{lian2023mer} across audio, visual, and textual modalities. It is annotated by $11$ annotators through a human-LLM collaboration strategy, covering $236$ emotion categories with $1$ to $9$ labels per sample. The dataset does not constrain the label space, distinguishing it from conventional multimodal emotion recognition (MER) datasets that rely on predefined categories.

\textbf{EmoBench-M} \cite{emobench-m} is a benchmark for evaluating the emotional intelligence of multimodal large language models, covering three hierarchical levels—foundational emotion recognition, conversational emotion understanding, and socially complex emotion analysis—across $13$ scenarios. It employs video and audio data for tasks including speech and music emotion recognition, opinion sentiment analysis, multi-party dialogue emotion recognition, humor understanding, and sarcasm detection. Each sample undergoes multi-annotator review, with accuracy for classification and model-based evaluation for generation.

\textbf{MER-Caption} \cite{AffectGPT} is a large-scale emotion description dataset comprising approximately $115K$ coarse-grained and $31K$ fine-grained labeled samples, covering over $2,000$ fine-grained emotion categories. The dataset is constructed under a model-led, human-assisted annotation framework, incorporating human priors to guide description generation and sample filtering, thereby balancing label quality with dataset scale. The raw data is sourced from the unlabeled portion of MER2024~\cite{lian2024mer}, spanning audio, visual, and textual modalities.

\textbf{EmoReAlM} \cite{AVERE} is a benchmark for evaluating audio-visual emotion reasoning in MLLMs, comprising approximately $4,000$ human-verified multiple-choice questions across $3$ task categories: basic reasoning, modality agreement, and stress testing. It investigates whether models rely on irrelevant cues or fabricate audio-visual evidence for emotion inference, thereby assessing their reasoning reliability. Samples are sourced from DFEW and formatted as multiple-choice questions to support reproducible evaluation.

\textbf{LIRIS-ACCEDE} \cite{LIRIS-ACCEDE} is a video database for affective content analysis, comprising $9,800$ clips from $160$ films, each lasting $8$ to $12$ seconds and spanning diverse genres. Annotations were obtained via crowdsourced pairwise comparisons, yielding relative rankings along the induced valence and arousal dimensions. Annotators from $89$ countries participated with consistent inter-annotator agreement. Visual and audio features, along with four experimental protocols, are provided with the dataset.

\textbf{MediaEval} \cite{MediaEval} is an evaluation task based on the LIRIS-ACCEDE~\cite{LIRIS-ACCEDE} dataset, comprising regression-based prediction of valence and arousal levels, as well as binary classification of fear, at the granularity of $10$-second video segments. The development set includes $30$ films and the test set $14$ films. Valence and arousal were continuously annotated by 16 participants, while fear labels were provided by two experts indicating whether each segment is expected to elicit fear.

\textbf{VAD Danmu} \cite{vad-danmu} is a large-scale dataset for video affective content analysis, comprising $19,267$ segmented clips from user-generated videos on BiliBili. It is annotated via crowdsourcing with discrete valence and arousal labels, primary emotion labels, and pairwise comparisons between consecutive clips. Unlike existing video-only datasets, VAD Danmu additionally incorporates danmu—real-time user comments during video playback—providing supplementary semantic cues for affective analysis.

\begin{figure*}[] 
    \centering
    \includegraphics[width=\textwidth]{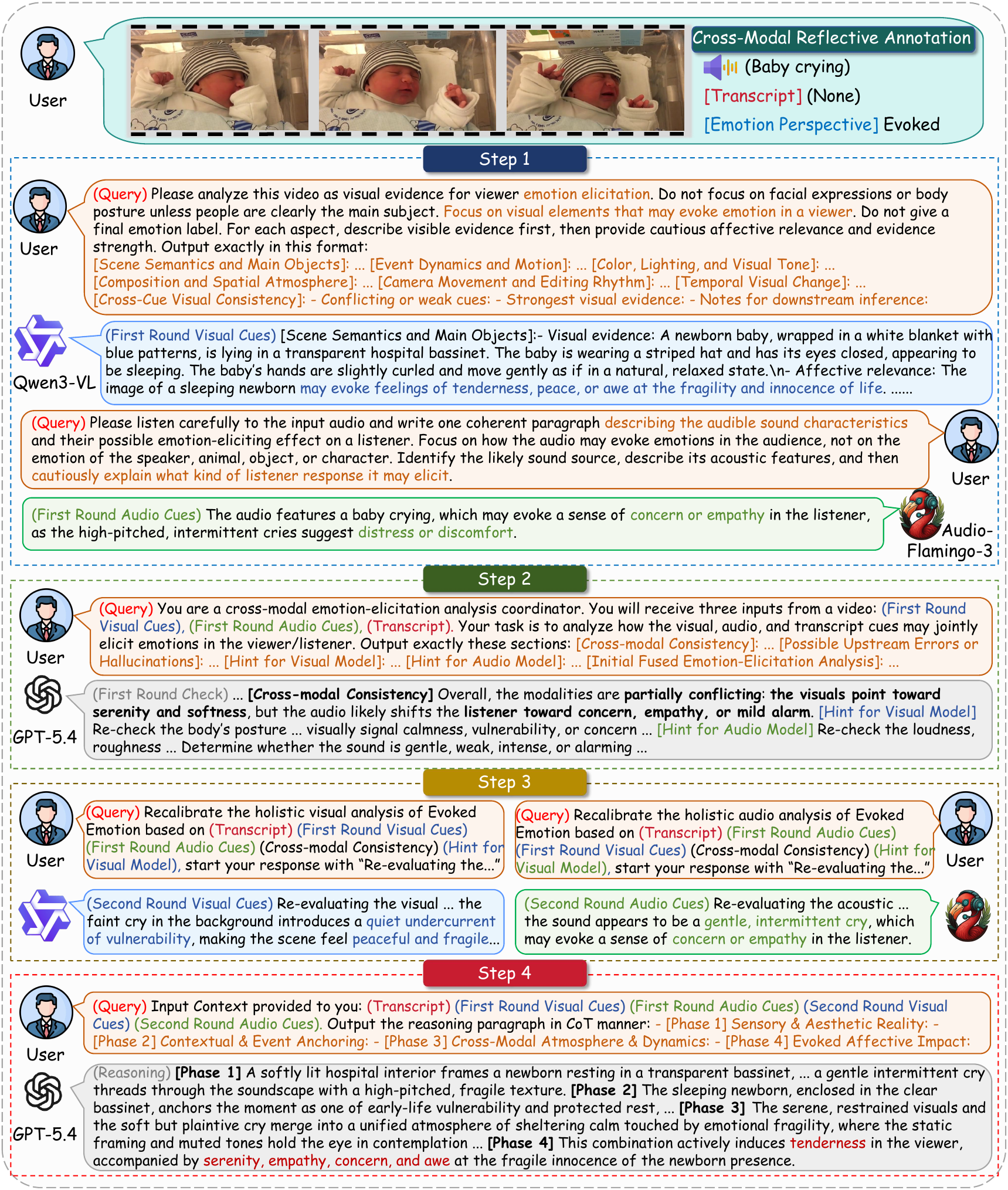}
    \caption{Detailed workflow of Cross-Modal Reflective Annotation, illustrated using evoked emotion as an example. The process comprises four stages, with the corresponding prompts and key model outputs shown for each stage.}
    \label{fig: CMRA prompt}
\end{figure*}

\textbf{EEmo-Bench} \cite{eemo-bench} is a benchmark dataset for evaluating multimodal large language models on image-evoked emotion assessment, comprising $1,960$ images collected from Flickr across diverse content categories. Each image is annotated with the top $3$ evoked emotions based on Ekman's seven basic categories plus neutral, along with three-dimensional VAD scores. The benchmark consists of $6,773$ question-answer pairs across four tasks: perception, ranking, description, and assessment.

\textbf{AICA-Bench} \cite{AICA-Bench} is a benchmark for evaluating vision-language models on affective image content analysis, covering $3$ core tasks: emotion understanding, reasoning, and emotion-guided generation. It integrates $9$ public datasets with $8,086$ images spanning social scenes, abstract art, and artistic photographs, annotated under Ekman~\cite{Ekman}, Mikels~\cite{Mikels}, and Plutchik~\cite{plutchik1980emotion} emotion models. The benchmark comprises $18,124$ standardized instructions for systematic evaluation across affective perception, causal reasoning, and emotionally aligned generation.

\section{Benchmark Construction Details} \label{sec: construction}

\subsection{Prompts for Emotion Evidence Bootstrapping}

Fig.~\ref{fig: CMRA prompt} illustrates the detailed workflow of Cross-Modal Reflective Annotation (CMRA) for emotion-evidence bootstrapping, using evoked emotion as an example. It presents the prompt and representative model output at each stage. CMRA first employs modality-specific MLLMs to extract affective cues from the visual and auditory streams, thereby transforming complex multimodal understanding into a text-based evidence-integration problem. A strong general-purpose model, such as \textit{GPT-5.4}~\cite{gpt5.4}, then attributes, organizes, and evaluates the extracted evidence. During reflective annotation, each modality-specific model revisits its initial analysis in light of \textit{GPT-5.4}’s feedback and complementary evidence from the other modality, recovering overlooked details and contextual information. Finally, \textit{GPT-5.4} integrates the revised evidence into a structured four-stage reasoning process that identifies affective causes and infers OV emotion concepts. Although the overall workflow is shared, the prompts are tailored to the target perspective. Evoked-emotion annotation emphasizes scene semantics, composition, spatial atmosphere, color, lighting, and visual tone, whereas expressed-emotion annotation focuses on facial expression, body movements, and posture.

\subsection{Annotation Process of Golden Anchors}

\begin{figure*}[] 
    \centering
    \includegraphics[width=0.95\textwidth]{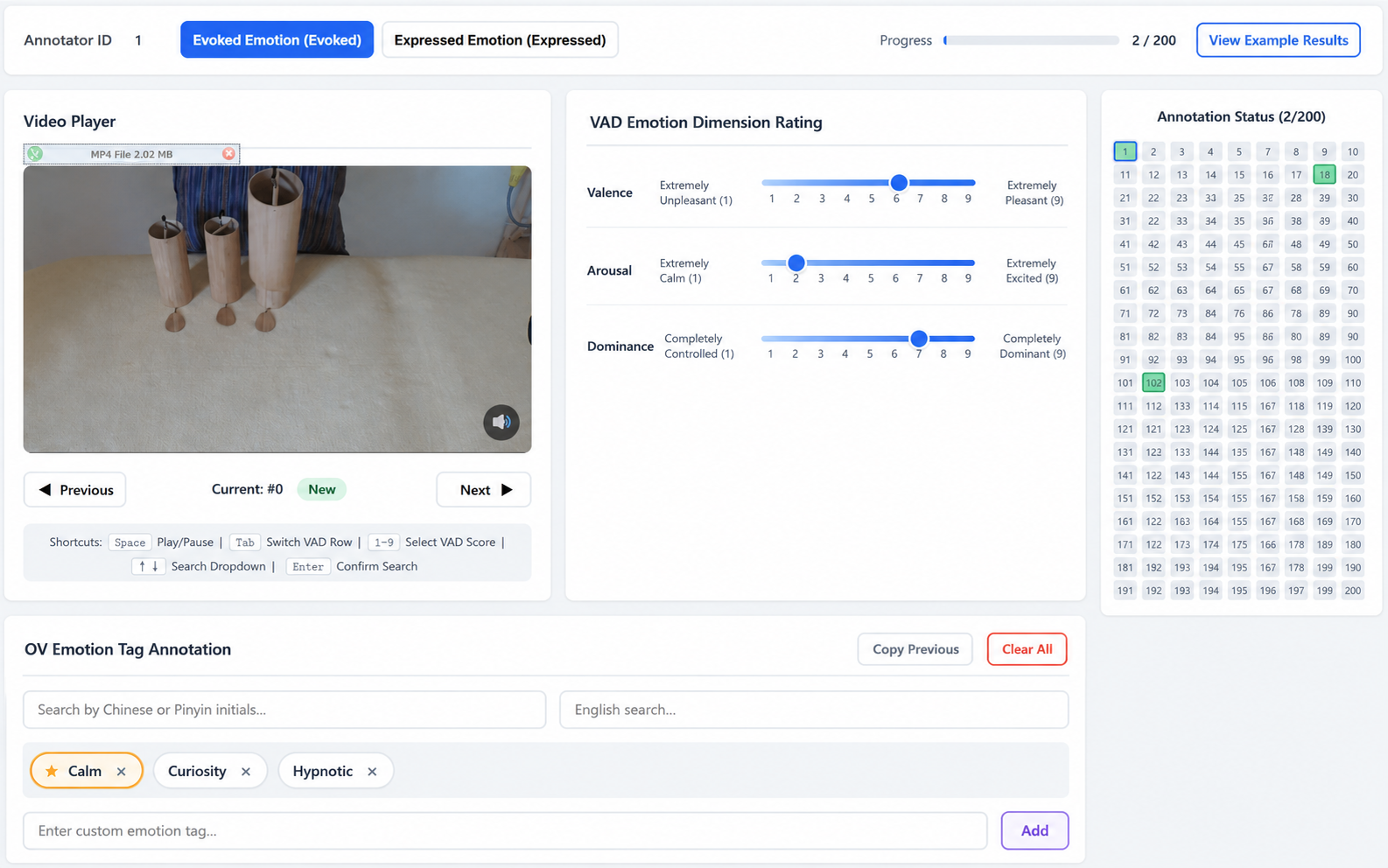}
    \caption{GUI for VAD annotation of the Golden Anchor Set of E$^3$mo-Bench.}
    \label{fig: round 1 GUI VAD}
\end{figure*}

\begin{figure*}[] 
    \centering
    \includegraphics[width=0.95\textwidth]{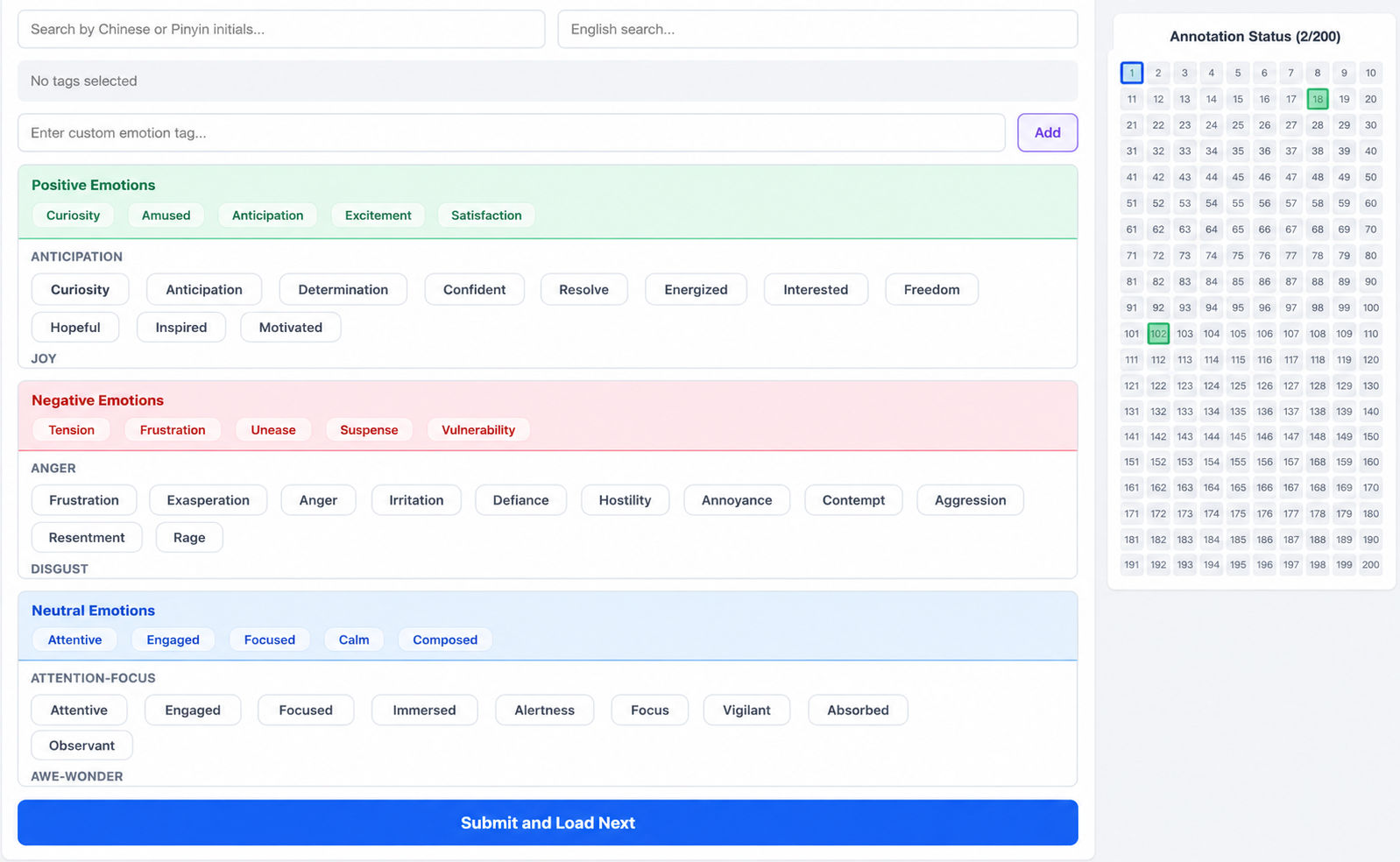}
    \caption{GUI for OV emotion annotation of the Golden Anchor Set of E$^3$mo-Bench.}
    \label{fig: round 1 GUI VAD-2}
\end{figure*}

Figs.~\ref{fig: round 1 GUI VAD} and ~\ref{fig: round 1 GUI VAD-2} illustrate the valence, arousal, and dominance (VAD) rating and OV-emotion annotation graphical user interfaces (GUIs) used for the Golden Anchor Set. Before annotation, all $15$ annotators receive conceptual instruction and complete example-based practice sessions. To mitigate fatigue, each annotator labels no more than $50$ videos per day. For each sample, annotators first watch the complete video with audio before providing any response. They then assign integer scores from $1$ to $9$ for VAD using $3$ independent sliders. Next, they select one or more emotion terms from a searchable hierarchical vocabulary of $130$ labels and may add a custom term when no predefined label adequately describes the emotion. One selected term is designated as the dominant emotion. Annotators evaluate each video independently based on their immediate impression while considering all available visual and auditory cues. Before submission, the interface verifies that all three VAD ratings, at least one emotion label, and one dominant emotion have been specified.

\subsection{VAD Data Processing and Validation of Golden Anchors}

The VAD annotation data are processed as follows: Let $\mathcal{H}=\{\mathrm{V},\mathrm{A},\mathrm{D}\}$ and let $r_{ijh}\in[1,9]$ denote the raw rating assigned by annotator $i$ to video $j$ on dimension $h\in\mathcal{H}$. Normalization, cleaning, and reliability analysis are performed separately for each dimension, followed by a common rescaling to the original rating range.

\paragraph{Annotator-wise Normalization.}
To reduce systematic differences in scale usage, the ratings of each annotator are standardized separately within each dimension:
\begin{equation}
    u_{ijh}=\frac{r_{ijh}-\bar r_{i h}}{s_{ih}},
    \qquad
    \bar r_{i h}=\frac{1}{N_i}\sum_{j=1}^{N_i}r_{ijh},
    \label{eq:ga_normalization}
\end{equation}
\begin{equation}
    s_{ih}=\sqrt{\frac{1}{N_i}\sum_{j=1}^{N_i}
    \left(r_{ijh}-\bar r_{i h}\right)^2},
    \label{eq:ga_annotator_std}
\end{equation}
where $N_i$ is the number of videos rated by annotator $i$. If $s_{ih}=0$, all corresponding normalized ratings are set to zero because they contain no information about differences among videos.

\paragraph{Outlier Rejection and Consensus Aggregation.}
For each video--dimension pair, we first compute the mean and standard deviation of the normalized ratings:
\begin{equation}
\begin{aligned}
    \bar u_{jh}&=\frac{1}{R_{jh}}\sum_{i}u_{ijh},\\
    t_{jh}&=\sqrt{\frac{1}{R_{jh}}\sum_{i}
    \left(u_{ijh}-\bar u_{jh}\right)^2}.
\end{aligned}
    \label{eq:ga_initial_statistics}
\end{equation}
where $R_{jh}$ is the number of available ratings. Following the adopted BT.500-style rule~\cite{ITU-R}, a rating is removed when
\begin{equation}
    \left|u_{ijh}-\bar u_{jh}\right|>1.96\,t_{jh}.
    \label{eq:ga_outlier}
\end{equation}
Let $\mathcal{A}_{jh}$ denote the retained annotator set and $c_{jh}=|\mathcal{A}_{jh}|$. The cleaned consensus statistics are then recomputed as
\begin{equation}
\begin{aligned}
    \widetilde u_{jh}&=\frac{1}{c_{jh}}
    \sum_{i\in\mathcal{A}_{jh}}u_{ijh},\\
    \widetilde t_{jh}&=\sqrt{\frac{1}{c_{jh}}
    \sum_{i\in\mathcal{A}_{jh}}
    \left(u_{ijh}-\widetilde u_{jh}\right)^2}.
\end{aligned}
    \label{eq:ga_clean_statistics}
\end{equation}

\paragraph{Rescaling to the Original Rating Range.}
Let $\bar r_{\mathrm{all}}$ and $s_{\mathrm{all}}$ be the mean and standard deviation computed over all raw VAD ratings, with $T$ denoting their total number:
\begin{equation}
\begin{aligned}
    \bar r_{\mathrm{all}}&=\frac{1}{T}\sum_{i,j,h}r_{ijh},\\
    s_{\mathrm{all}}&=\sqrt{\frac{1}{T}\sum_{i,j,h}
    \left(r_{ijh}-\bar r_{\mathrm{all}}\right)^2}.
\end{aligned}
    \label{eq:ga_global_statistics}
\end{equation}
The Golden Anchor rating is obtained by mapping the cleaned normalized mean back to the $1$--$9$ scale:
\begin{equation}
    \widehat r^{\mathrm{GA}}_{jh}
    =\operatorname{clip}\!\left(
    \widetilde u_{jh}s_{\mathrm{all}}+\bar r_{\mathrm{all}},\,1,\,9
    \right),
    \label{eq:ga_rescaling}
\end{equation}
where $\operatorname{clip}(x,\ell,u)=\max(\ell,\min(u,x))$.

\paragraph{Anchor Retention.}
A video is retained as a Golden Anchor only when at least $10$ valid ratings remain, and the cleaned normalized dispersion is below $1.5$ for every VAD dimension:
\begin{equation}
    b_j=\prod_{h\in\mathcal{H}}
    \mathbf{1}\!\left[c_{jh}\geq 10\ \wedge\ \widetilde t_{jh}<1.5\right].
    \label{eq:ga_retention}
\end{equation}
Videos with $b_j=1$ are retained, yielding 383 Golden Anchors from the 400 candidates.

\paragraph{inter-annotator reliability.}
Reliability is assessed separately for valence, arousal, and dominance using the raw rating matrices before normalization and outlier rejection. Because the same fixed panel of annotators rates the samples and the benchmark uses their average rating, we report the two-way mixed-effects, average-measures consistency coefficient ICC$(3,k)$~\cite{ICC}:
\begin{equation}
    \operatorname{ICC}(3,k)_h
    =\frac{MS_{\mathrm{video},h}-MS_{\mathrm{error},h}}
    {MS_{\mathrm{video},h}},
    \label{eq:ga_icc}
\end{equation}
where $MS_{\mathrm{video},h}$ and $MS_{\mathrm{error},h}$ are respectively the between-video and residual mean squares obtained from a two-way mixed-effects ANOVA for dimension $h$, and $k$ is the number of annotators. Higher values indicate greater consistency of the average human rating.

We assess the inter-annotator reliability of the Golden Anchor VAD annotations using ICC$(3,k)$, obtaining values of $0.89$, $0.73$, and $0.56$ for valence, arousal, and dominance, respectively. These results indicate moderate-to-good consistency of the averaged human ratings in the inherently subjective setting of affective annotation.

\subsection{OV Emotion Processing of Golden Anchors}

Human OV annotations may contain synonymous
expressions, repeated concepts, and low-frequency noise. We first
normalize all emotion terms to a canonical vocabulary and merge the
votes of synonymous terms. The dominant emotion is selected by
majority vote. If several candidates receive the same highest vote
count, the tie is resolved using their centrality within the
sample-specific OV distribution in VAD space. Let
$\mathcal{D}_{i}$ denote the tied candidates for video $i$,
$v_{i}(y)$ denote the aggregated OV vote count of emotion $y$, and
$\boldsymbol{\phi}(y)=(V(y),A(y),D(y))^{\mathsf T}$ denote the VAD
vector of emotion $y$
The final dominant emotion is
\begin{equation}
    e_{i}^{\star}
    =\arg\min_{x\in\mathcal{D}_{i}}
    \frac{\sum_{y}v_{i}(y)
    \left\lVert\boldsymbol{\phi}(x)-\boldsymbol{\phi}(y)\right\rVert_{2}}
    {\sum_{y}v_{i}(y)}.
    \label{eq:ga_ov_dominant}
\end{equation}
Thus, the selected candidate is the one closest to the overall
affective profile of the sample.

To construct a compact OV label set, we retain canonical emotions
receiving at least two votes as core labels. If at least four core
labels are available, they are ranked by vote count and truncated to
at most five terms. Otherwise, additional low-vote labels are drawn
from semantic clusters not yet represented by the core set. Within
each uncovered cluster, the representative is selected by prioritizing
higher sample-level votes, greater corpus-level frequency, and smaller
VAD distance to the dominant emotion, in that order. Cluster
representatives are then added according to their total vote support
until four labels are obtained, while the final set remains capped at
five. This procedure preserves the principal affective content while
limiting synonym redundancy and long-tail noise.

\begin{figure*}[t] 
    \centering
    \includegraphics[width=0.95\textwidth]{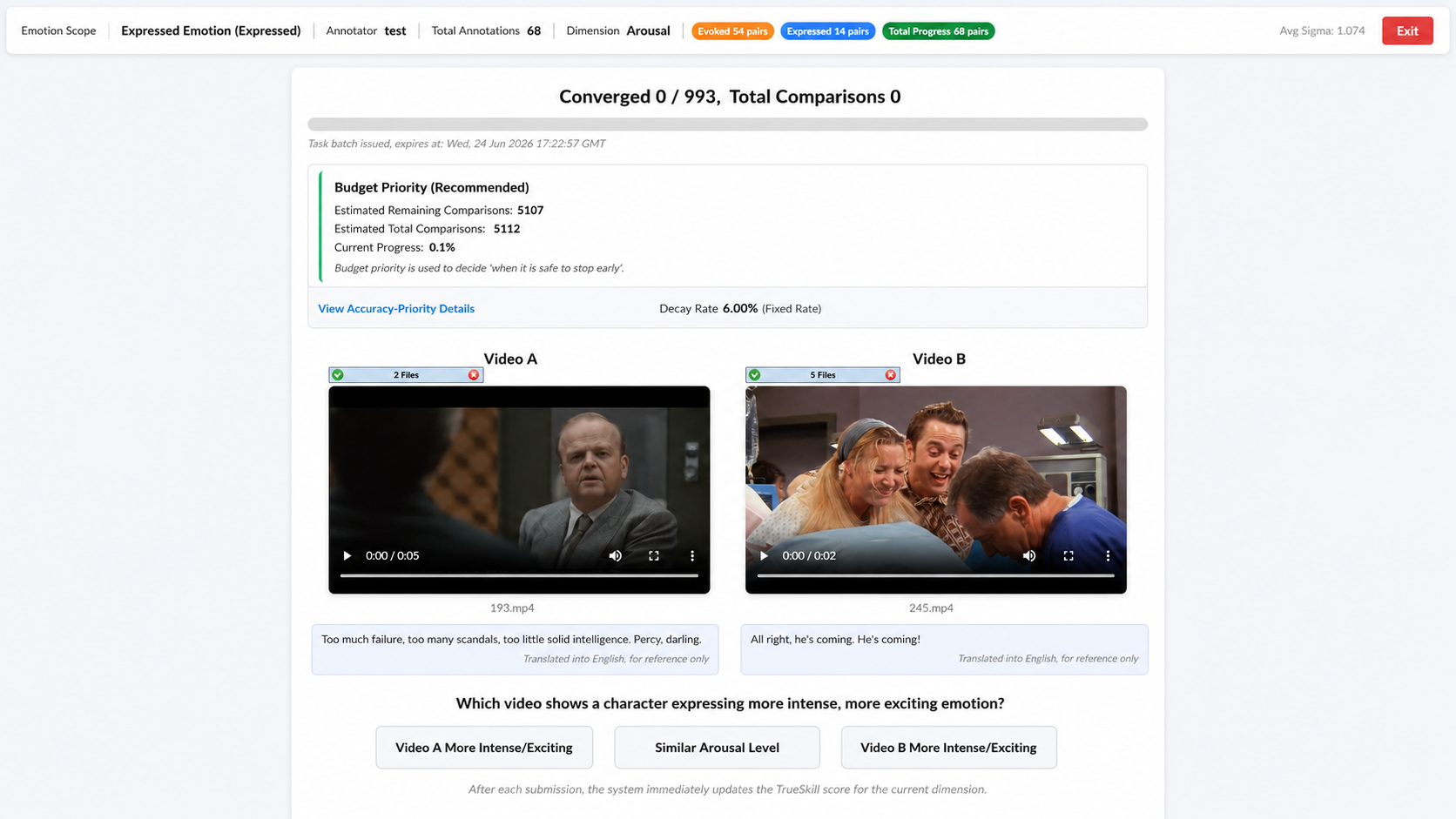}
    \caption{GUI for Human Pairwise VAD Annotation of the Broad-Coverage Set of E$^3$mo-Bench.}
    \label{fig: round 2 GUI VAD}
\end{figure*}

\subsection{Protocol for Difficulty-Stratified Sampling within the Broad-Coverage Set}

This section details the difficulty-stratified sampling strategy used to construct the Broad-Coverage Evaluation Set. Because \textbf{medium}-difficulty samples are added after selecting the \textbf{easy} and \textbf{hard} subsets to balance the distribution of preliminary VAD estimates, we focus here on the selection criteria for easy and hard samples.

\subsubsection{Easy-Sample Selection.}

We estimate the within-sample affective concentration of the
preliminary OV emotion labels in the VAD space. For each candidate
video $i$, we retain only the emotion terms matched to the strict VAD
lexicon. Let $\mathcal{E}_i$ denote the resulting set, with
$n_i=|\mathcal{E}_i|\geq 2$. We compute the centroid and weighted dispersion
as
\begin{equation}
\begin{aligned}
\bar{\boldsymbol{\phi}}_i
&=
\frac{1}{n_i}
\sum_{e\in\mathcal{E}_i}\boldsymbol{\phi}(e),\\
D_i^{\mathrm{VAD}}
&=
\frac{1}{n_i}
\sum_{e\in\mathcal{E}_i}
\left(
\boldsymbol{\phi}(e)-\bar{\boldsymbol{\phi}}_i
\right)^{\mathsf T}
\mathbf{W}
\left(
\boldsymbol{\phi}(e)-\bar{\boldsymbol{\phi}}_i
\right),
\end{aligned}
\label{eq:easy_vad_dispersion}
\end{equation}
where
$\mathbf{W}=\operatorname{diag}(\omega_v,\omega_a,\omega_d)$,
$\omega_v,\omega_a,\omega_d\geq 0$, and
$\omega_v+\omega_a+\omega_d=1$. In our implementation, we set
$(\omega_v,\omega_a,\omega_d)=(0.50,0.35,0.15)$ to emphasize
affective polarity and activation while retaining dominance as a
complementary component.

A lower $D_i^{\mathrm{VAD}}$ indicates that the OV emotion terms are
more concentrated around a common affective center. Samples with fewer
than two matched terms are excluded. Among the candidates passing the preceding relaxed quality filtering, we rank the samples by $D_i^{\mathrm{VAD}}$ in ascending order, selecting the first $50\%$ of the Broad Coverage Set to form the easy subset. This measure is used only for difficulty
stratification, rather than as an evaluation metric.

\subsubsection{Hard-Sample Selection.}

We construct the hard subset by identifying samples that require
nontrivial integration of heterogeneous audio-visual evidence. We
first use the \textbf{cross-modal consistency} analysis produced during the
first GPT-based fusion stage of CMRA to retrieve candidates exhibiting
pronounced disagreement between auditory and visual affective cues.
Because such disagreement may also arise from noisy or implausible
modality analyses, it is not treated as sufficient evidence of sample
difficulty. \textit{GPT-5.4} subsequently examines the final OV emotion rationale
to determine whether the two modalities are integrated in a coherent,
self-consistent, and contextually plausible manner.

Candidates passing this plausibility check are rated on a five-point
scale along $3$ complementary criteria: 1) \emph{Asymmetry} measures
the extent to which the auditory and visual modalities support
different or conflicting affective interpretations. 2) \emph{Synergy}
measures how strongly the final emotion interpretation depends on
joint evidence from both modalities rather than either modality alone.
3) \emph{Grounding} measures whether the integrated rationale is
supported by identifiable audio-visual evidence. Samples receiving
consistently high scores across these criteria are prioritized for the
hard subset, as they exhibit complex yet interpretable cross-modal
affective structures rather than arbitrary modality disagreement.

\subsection{Protocol and Quality Validation for Human Pairwise VAD Annotation}

\subsubsection{Annotation Process.}
The annotation GUI is illustrated in Fig.~\ref{fig: round 2 GUI VAD}.
Before annotation, participants receive standardized instructions and
examples clarifying the $3$ VAD dimensions and the distinction
between evoked and expressed emotion. The overall annotation pipeline is divided into $6$ distinct subtasks. To minimize cross-dimensional interference, each annotation session is strictly confined to a single affective perspective and one specific VAD dimension. For the evoked-emotion pool,
annotators compare the affective responses induced in themselves by
the two videos, whereas for the expressed-emotion pool, they compare
the emotions conveyed by the depicted subjects. Annotators first watch
both videos in full, with repeated playback and subtitles permitted
only as auxiliary information, and then judge exclusively along the
assigned dimension: emotional positivity for valence, activation
intensity for arousal, or perceived control and submissiveness for
dominance. Each comparison is recorded as Video A being higher, the
two videos being similar, or Video B being higher. The similar option
is selected only when no meaningful difference remains after careful
inspection. Annotators use a fixed identifier throughout the study and
apply a consistent personal criterion across comparisons. Initial
calibration items and periodically inserted anchor comparisons are
completed in the same manner as ordinary trials and are used to verify
task understanding and maintain a stable annotation scale. 
For each subtask, multiple annotators collaboratively evaluate the dispatched pairwise comparisons. The pair-dispatching process dynamically halts, and the final scores are derived once the mean uncertainty $\sigma$ across all sample scores drops below a predefined threshold $\sigma_\tau$.

\begin{figure*}[] 
    \centering
    \includegraphics[width=0.75\textwidth]{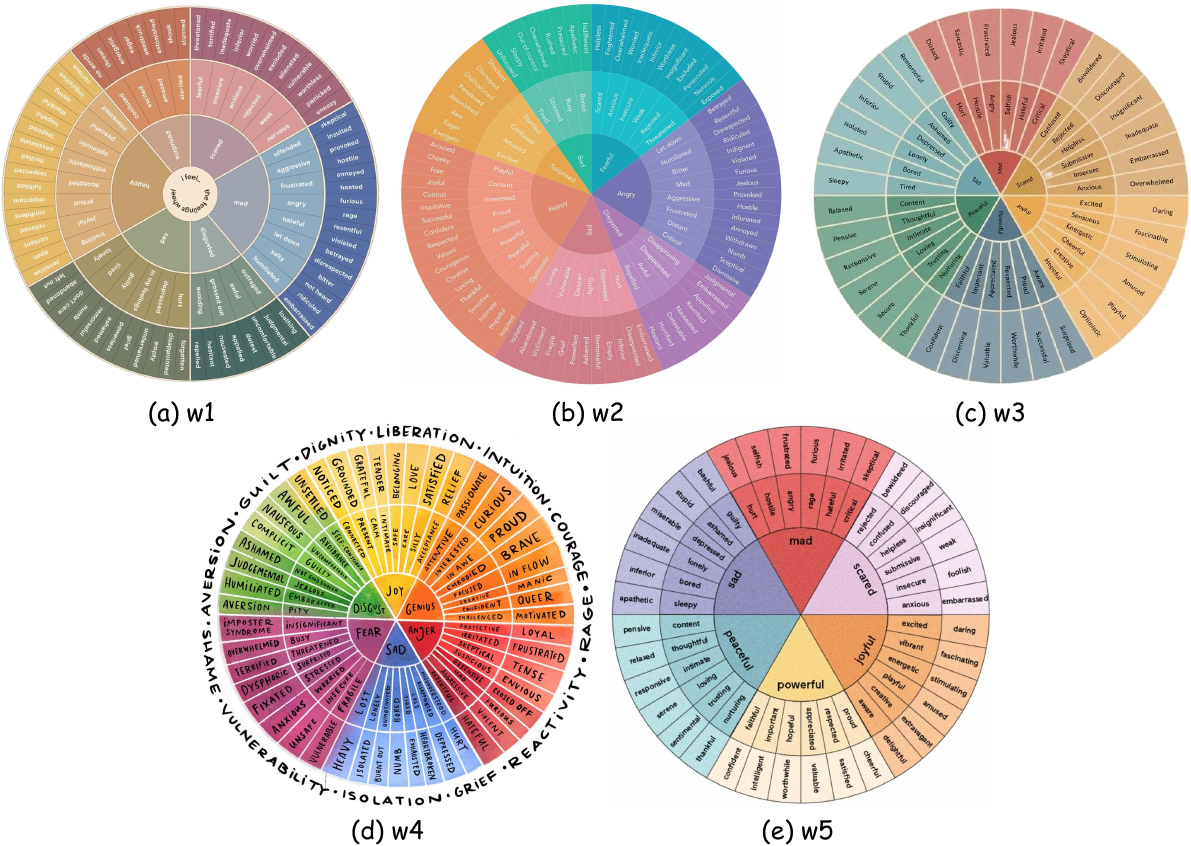}
    \caption{Emotion wheels formulated following the OV-MER settings~\cite{OV-MER}. This figure illustrates the $5$ distinct emotion wheels utilized in our framework for categorical affective representation.}
    \label{fig: emotion wheel}
\end{figure*}

\subsubsection{Annotation Quality Validation.}



To assess the quality of human pairwise VAD annotation, we conduct a small-scale validation on the expressed-valence dimension using Golden Anchor samples with established mean opinion score (MOS) scores. Following the E$^3$mo-Score protocol, approximately $5\%$ of the samples retain their reference scores as fixed anchors, while the remaining scores are masked and treated as candidates. Human annotators judge the sample pairs selected by the original online scheduling strategy, and BPA periodically refines the candidate scores through global Bayesian optimization. Optimization terminates once the mean uncertainty across candidate samples falls below the predefined threshold $\sigma_{\tau}$. The inferred scores achieve a Spearman Rank-Order Correlation Coefficient (SRCC) of $0.91$ and a Pearson Linear Correlation Coefficient (PLCC) of $0.90$ against the reference MOS scores, indicating that BPA-supported pairwise annotation can recover highly consistent continuous ratings and providing evidence for the reliability of the VAD scores in the Broad-Coverage Set.

\subsection{Cost Comparison of VAD Annotation Paradigms}

We estimate annotation cost by accounting for both data validity and task-specific effort. In the Golden Anchor annotation, 383 of 400 samples pass the consistency-based quality control, corresponding to a valid-sample rate of $95.75\%$. Under the same retention rate, obtaining 2,141 valid samples through conventional MOS annotation would require approximately $2{,}141/0.9575\approx2{,}236$ initially annotated samples, or about $33{,}540$ complete rating sessions with 15 annotators per sample. This workload is comparable to the 35,639 pairwise judgments collected by BPA. Based on pilot completion times and the resulting compensation scheme, each pairwise judgment incurs approximately $60\%$ of the cost of a complete MOS session because it requires a simpler relative decision without maintaining an absolute rating scale. Consequently, the estimated total cost of BPA is approximately $64\%$ of conventional MOS annotation, or roughly $60\%$ when reported at a coarse level.

\section{Supplementary Methodological Details} \label{sec: algorithm}

\subsection{Semantic Emotion-Wheel Alignment}
\label{sec:emotion_wheel_metric}

We evaluate OV emotion predictions using a sample-wise
adaptation of the emotion-wheel alignment metric introduced in
AffectGPT~\cite{AffectGPT}. Before scoring, predicted and reference
emotion terms are lowercased, deduplicated, and mapped to a canonical
emotion vocabulary through morphological normalization and synonym
resolution. We then project each canonical emotion onto $5$ established emotion
taxonomies: \textit{Plutchik's emotion wheel}~\cite{plutchik1980emotion},
\textit{Parrott's hierarchical taxonomy}~\cite{parrott2001emotions},
\textit{Shaver's prototype taxonomy}~\cite{shaver1987emotion},
\textit{the Geneva Emotion Wheel}~\cite{scherer2013grid}, and \textit{Russell's
circumplex model}~\cite{russell1980circumplex}.
Following OV-MER~\cite{OV-MER}, we operationalize these
taxonomies as $5$ canonical emotion-to-category mappings (illustrated in Fig.~\ref{fig: emotion wheel}). 

Let $\mathcal{R}_i$ and $\mathcal{P}_i$ denote the reference and
predicted canonical emotion sets for sample $i$, respectively. For the
$k$-th emotion wheel, let $\Gamma_k(\cdot)$ map an emotion set to the
set of wheel sectors occupied by its elements. The wheel-specific
precision, recall, and F-score are defined as
\begin{equation}
\begin{aligned}
P_{ik}^{\mathrm{EW}}
&=
\frac{
\left|
\Gamma_k(\mathcal{R}_i)
\cap
\Gamma_k(\mathcal{P}_i)
\right|
}{
\left|
\Gamma_k(\mathcal{P}_i)
\right|
},\\
R_{ik}^{\mathrm{EW}}
&=
\frac{
\left|
\Gamma_k(\mathcal{R}_i)
\cap
\Gamma_k(\mathcal{P}_i)
\right|
}{
\left|
\Gamma_k(\mathcal{R}_i)
\right|
},\\
F_{ik}^{\mathrm{EW}}
&=
\frac{
2P_{ik}^{\mathrm{EW}}R_{ik}^{\mathrm{EW}}
}{
P_{ik}^{\mathrm{EW}}+R_{ik}^{\mathrm{EW}}
}.
\end{aligned}
\label{eq:ew_per_wheel}
\end{equation}

The sample-level score averages the results over the $K=5$ emotion
wheels, and the benchmark score is the macro-average over all $N$
samples:
\begin{equation}
F_i^{\mathrm{EW}}
=
\frac{1}{K}\sum_{k=1}^{K}F_{ik}^{\mathrm{EW}},
\qquad
F^{\mathrm{EW}}
=
\frac{1}{N}\sum_{i=1}^{N}F_i^{\mathrm{EW}}.
\label{eq:ew_overall}
\end{equation}

Only exact sector-level overlap is counted. Emotion frequencies are
not used, and no partial credit is assigned to adjacent sectors. By
aggregating multiple emotion-wheel taxonomies, the metric reduces
sensitivity to any single categorical organization of the emotion
space.

\subsection{TrueSkill-Based Provisional Updates}
\label{sec:trueskill_update}

TrueSkill~\cite{trueskill} is used to provisionally update
sample scores and uncertainties after each pairwise judgment. These
online estimates guide subsequent pair selection, whereas the final
scores are recomputed by the periodic global BPA optimization using all
collected judgments. To distinguish the provisional TrueSkill state
from the final BPA variables, we denote the score and uncertainty of
sample $i$ by $m_i$ and $u_i$, respectively.

TrueSkill represents the latent score of each sample as
$x_i\sim\mathcal{N}(m_i,u_i^2)$. Its observed performance is modeled as
$r_i\sim\mathcal{N}(x_i,\beta_{\mathrm{TS}}^2)$, where $\beta_{\mathrm{TS}}^2$ accounts for
comparison noise. For a pair $(i,j)$, the performance difference
therefore follows

\begin{equation}
    r_i-r_j
\sim
\mathcal{N}\!\left(\Delta_{ij},c_{ij}^{2}\right),
\Delta_{ij}=m_i-m_j,
c_{ij}^{2}=u_i^{2}+u_j^{2}+2\beta_{\mathrm{TS}}^{2},
\end{equation}

\noindent where $c_{ij}$ is the predictive standard deviation of the pairwise performance difference, combining the current uncertainties of both samples with the observation noise of the comparison. Let $\epsilon>0$ denote the draw margin and $\Phi(\cdot)$ the standard
normal cumulative distribution function. The probabilities of sample
$i$ being higher, the two samples being similar, and sample $j$ being
higher are

\begin{equation}
    \begin{aligned}
Z_{+}
&=
\Phi\!\left(\frac{\Delta_{ij}-\epsilon_{\mathrm{draw}}}{c_{ij}}\right),\\
Z_{0}
&=
\Phi\!\left(\frac{\epsilon_{\mathrm{draw}}-\Delta_{ij}}{c_{ij}}\right)
-
\Phi\!\left(\frac{-\epsilon_{\mathrm{draw}}-\Delta_{ij}}{c_{ij}}\right),\\
Z_{-}
&=
\Phi\!\left(\frac{-\Delta_{ij}-\epsilon_{\mathrm{draw}}}{c_{ij}}\right).
\end{aligned}
\end{equation}

\begin{equation}
    \epsilon_{\mathrm{draw}}
=
\sqrt{2}\,\beta_{\mathrm{TS}}
\Phi^{-1}
\left(
\frac{1+p_{\mathrm{draw}}}{2}
\right).
\end{equation}

\noindent For the observed outcome $y\in\{+,0,-\}$, let $Z_y$ denote its
predictive probability defined above. We define the first-order score
correction and local observed information as
\begin{equation}
g_y
=
\frac{\partial \log Z_y}{\partial \Delta_{ij}},
\qquad
J_y
=
-\frac{\partial^2 \log Z_y}
{\partial \Delta_{ij}^{2}}.
\label{eq:trueskill_correction}
\end{equation}

\noindent Here, $g_y$ determines the direction and magnitude of the score update
implied by the observed comparison, whereas $J_y$ measures the local
curvature of the outcome likelihood and controls the corresponding
reduction in uncertainty. A more informative outcome generally yields
a larger $J_y$. Gaussian moment matching gives the provisional posterior updates

\begin{equation}
\begin{aligned}
m_i' &= m_i + u_i^2 g_y,
&
m_j' &= m_j - u_j^2 g_y,\\
{u_i'}^{2} &= u_i^2 - u_i^4 J_y,
&
{u_j'}^{2} &= u_j^2 - u_j^4 J_y.
\end{aligned}
\label{eq:trueskill_update}
\end{equation}
\noindent The sign of $g_y$ moves the two score estimates toward the ordering
supported by the observed outcome; for a similar judgment, it instead
moves their difference toward the draw region. Meanwhile, $J_y$
reduces the posterior variances according to the information provided
by the comparison. These updates are used only as provisional online
states for pair scheduling, while the final scores are recomputed by
the periodic global BPA optimization.

Specifically, the same provisional TrueSkill update is applied to all dispatched
pairs, regardless of the pair-selection strategy. Exploration selects
pairs according to the current score proximity and uncertainty,
whereas \textbf{Anchor Calibration} and \textbf{Bridge Sampling} follow their respective
routing criteria. For an anchor-calibration pair, both the candidate
and anchor are temporarily updated using the observed comparison.
Immediately afterward, the anchor state is restored to its fixed
reference values $(\mu_\mathcal{G}^{(0)},\sigma_\mathcal{G}^{(0)})$, preventing the calibration
scale from drifting while retaining the update induced on the
candidate. For a bridge pair, both candidate samples retain their
updated provisional states. \textbf{Thus, the dispatch strategies differ only
in how comparison pairs are selected, while all online state updates
follow the same TrueSkill inference rule.} These provisional states are
used solely for subsequent scheduling and are later replaced by the
periodic global BPA estimates.

\subsection{Percentile-Based Initialization of Candidate Samples in BPA}

The human pairwise annotation and E$^3$mo-Score pipelines use the same
initialization procedure for candidate samples. Each sample is
associated with a preliminary standardized VAD estimate obtained
during the preceding reference-annotation stage. These values are
used only to provide an ordered initialization and are subsequently
updated by pairwise evidence.

Because the preliminary VAD estimates are unbounded standardized
values, they cannot be used directly as scores in the internal
$[1,9]$ space. Direct use would also make initialization sensitive to
long-tailed values and could place samples prematurely near the score
boundaries. We therefore apply a perspective-specific and dimension-specific
percentile mapping that preserves the central ordering, clips extreme
values, and leaves sufficient room for subsequent correction by
pairwise evidence. This transformation serves only to calibrate the
initial prior scale rather than to define the final VAD labels.

Let $\zeta_{i,h}^{(t)}$ denote the preliminary standardized score of
sample $i$ for affective perspective
$t\in\{\mathrm{evoked},\mathrm{expressed}\}$ and VAD dimension
$h\in\{\mathrm{V},\mathrm{A},\mathrm{D}\}$. For each perspective and
dimension, we collect the scores of all candidate samples and compute
their $5$th and $95$th percentiles, denoted by
$\zeta_{t,h}^{-}$ and $\zeta_{t,h}^{+}$, respectively. We then define
the percentile-based mapping
\begin{equation}
\mathcal{M}_{t,h}(\zeta)
=
\operatorname{clip}_{[L,U]}
\left[
L+
(U-L)
\frac{\zeta-\zeta_{t,h}^{-}}
     {\zeta_{t,h}^{+}-\zeta_{t,h}^{-}}
\right],
\label{eq:non_anchor_mapping}
\end{equation}
where $\operatorname{clip}_{[L,U]}(\cdot)$ truncates its argument to
$[L,U]$. We set $L=2$ and $U=8$, such that the two percentile
boundaries are mapped to $2$ and $8$, while more extreme values are
clipped to the same interval.

The initial BPA state of a candidate sample is therefore
\begin{equation}
\begin{aligned}
    \mu_{i,h}^{(0)}
&=
\begin{cases}
\mathcal{M}_{t,h}\!\left(\zeta_{i,h}^{(t)}\right),
& \text{preliminary score is available},\\[2pt]
5,
& \text{otherwise},
\end{cases}
\\
\sigma_{i,h}^{(0)}&=\sigma_{\mathrm{init}},
\end{aligned}
\label{eq:non_anchor_initial_state}
\end{equation}
where $\sigma_{\mathrm{init}}=1.333$. The mapped value also serves as
the reference center of the adaptive prior in the subsequent global
maximum a posteriori (MAP) optimization. When a preliminary VAD estimate is unavailable, the
sample is initialized at the neutral midpoint and retains the same
initial uncertainty.

This percentile transformation is computed independently for each
affective perspective and VAD dimension. It reduces the influence of
extreme preliminary values, preserves their relative ordering over the
central data range, and avoids initializing candidate samples at the
absolute endpoints of the final $[1,9]$ scale. The resulting values
serve only as provisional prior locations; final scores are determined
by the accumulated pairwise judgments.

\subsection{Annotator Reliability Estimation}
\label{sec:annotator_reliability}

For human pairwise annotation, we estimate a separate reliability
coefficient for each annotator, affective perspective, and VAD
dimension. The perspective and dimension indices are omitted below
for clarity. Rather than optimizing the bounded coefficient
$\eta_k\in(0,1)$ directly, we introduce an unconstrained parameter
$\xi_k\in\mathbb{R}$ and define
\begin{equation}
    \eta_k
    =
    \operatorname{sigmoid}(\xi_k).
    \label{eq:eta_parameterization}
\end{equation}
All $\xi_k$ are initialized to zero, corresponding to an initial
reliability of $\eta_k=0.5$.

To regularize reliability estimates under sparse observations, we
place a Beta prior on each coefficient:
\begin{equation}
    \eta_k
    \sim
    \operatorname{Beta}(a_{\eta},b_{\eta}),
    \qquad
    a_{\eta}=10,\quad b_{\eta}=2.
    \label{eq:eta_beta_prior}
\end{equation}
Ignoring the normalization constant of the Beta density, the
corresponding negative log-prior is
\begin{equation}
    \mathcal{L}_{\eta}
    =
    -\lambda_{\eta}
    \sum_k
    \left[
        (a_{\eta}-1)\log\eta_k
        +
        (b_{\eta}-1)\log(1-\eta_k)
    \right],
    \label{eq:eta_prior_loss}
\end{equation}
where $\lambda_{\eta}$ controls the prior strength. This prior favors
relatively reliable, but non-deterministic, annotators and prevents
poorly supported estimates from collapsing toward either boundary.

The complete objective for human annotation is
\begin{equation}
    \mathcal{L}_{\mathrm{total}}^{'}
    =
    \mathcal{L}_{\mathrm{data}}
    +
    \mathcal{L}_{\mathrm{prior}}
    +
    \mathcal{L}_{\eta},
    \label{eq:human_bpa_objective}
\end{equation}
where $\mathcal{L}_{\mathrm{data}}$ is the tie-aware comparison
negative log-likelihood defined in the main text, and
$\mathcal{L}_{\mathrm{prior}}$ denotes the adaptive score-prior term.
Candidate scores, annotator reliabilities, and the shared tie margin
are jointly updated during each periodic global MAP optimization.

For completeness, let
$\mathcal{I}_k=\{m:k_m=k\}$ denote the indices of comparisons provided
by annotator $k$, and define
\begin{equation}
    \pi_m=p_{i_mj_m}^{y_m},
    \qquad
    \widetilde q_m
    =
    \eta_k\pi_m+\frac{1-\eta_k}{3},
\label{eq:eta_compact_probability}
\end{equation}
for $m\in\mathcal{I}_k$. Here, $\pi_m$ is the Rao--Kupper probability
assigned to the observed outcome, and $\widetilde q_m$ is the
corresponding reliability-adjusted probability. The gradient with
respect to the unconstrained reliability parameter is
\begin{equation}
\begin{aligned}
\frac{\partial\mathcal{L}_{\mathrm{total}}^{'}}{\partial\xi_k}
=&
\eta_k(1-\eta_k)
\Bigg[
-\sum_{m\in\mathcal{I}_k}
\frac{\pi_m-\frac{1}{3}}{\widetilde q_m}
\\
&-\lambda_\eta
\left(
\frac{a_\eta-1}{\eta_k}
-
\frac{b_\eta-1}{1-\eta_k}
\right)
\Bigg].
\end{aligned}
\label{eq:eta_gradient}
\end{equation}

The parameters $\{\xi_k\}$ are jointly optimized with the candidate
scores and tie margin during each periodic global MAP update.
Judgments that remain probable under the inferred global score
structure tend to increase $\eta_k$, whereas repeatedly conflicting
judgments tend to decrease it. Therefore, $\eta_k$ represents a
task-conditioned model estimate of annotator reliability rather than
an independent empirical accuracy measure.

\begin{table*}[t]
\centering
\renewcommand{\arraystretch}{1.08}
\setlength{\tabcolsep}{5pt}
\resizebox{\textwidth}{!}{
\begin{tabular}{l l c p{9.2cm}}
\hline
\textbf{Module} &
\textbf{Symbol} &
\textbf{Value} &
\textbf{Definition} \\
\hline

\multicolumn{4}{l}{\textit{Candidate samples initialization}}\\ \hdashline
Fallback score
& $\mu_{\mathrm{mid}}$
& $5.0$
& Neutral midpoint used when a candidate sample has no valid
preliminary VAD estimate. \\

Initial uncertainty
& $\sigma_{\mathrm{init}}$
& $1.333$
& Initial uncertainty assigned to every candidate sample before
online pairwise updating. \\

Mapping range
& $[L,U]$
& $[2,8]$
& Output interval of the percentile-based mapping from preliminary
standardized VAD estimates to initial BPA scores. \\

Mapping percentiles
& $(q_{\mathrm{lo}},q_{\mathrm{hi}})$
& $(5\%,95\%)$
& Lower and upper percentiles computed independently within each
affective-perspective--dimension pool for robust score initialization. \\

Anchor uncertainty
& $\sigma_{\mathcal{G}}$
& $10^{-4}$
& Near-zero uncertainty assigned to fixed Golden Anchors during
online scheduling. \\

\hdashline
\multicolumn{4}{l}{\textit{Pair Scheduling}}\\ \hdashline
Performance noise
& $\beta_{\mathrm{TS}}$
& $0.8$
& Standard deviation of the per-sample performance noise used in
the provisional TrueSkill update. \\

Draw probability
& $p_{\mathrm{draw}}$
& $0.15$
& Prior draw probability used to determine the TrueSkill draw
margin for similar judgments. \\

Bridge score-gap threshold
& $\theta$
& $3.0$
& Minimum absolute difference between the current provisional scores
of two samples considered for Bridge Sampling. \\

\hdashline
\multicolumn{4}{l}{\textit{Periodic global BPA optimization}}\\ \hdashline
Tie margin
& $\delta^{(0)}$
& $0.5$
& Initial value of the Rao--Kupper tie margin.  \\

Score-prior strength
& $\lambda_0$
& $1.0$
& Base strength of the adaptive quadratic prior centered at
$\mu_i^{(0)}$. \\

Prior decay
& $\gamma$
& $0.01$
& Rate at which the score prior weakens as the comparison count
$N_i$ increases. \\

Reliability prior
& $(a_{\eta},b_{\eta})$
& $(10,2)$
& Shape parameters of the Beta prior imposed on the task-conditioned
annotator reliability $\eta_k$. \\

Reliability regularization
& $\lambda_{\eta}$
& $1.0$
& Weight of the annotator-reliability negative log-prior in the human
BPA objective. \\

\hdashline
\multicolumn{4}{l}{\textit{Uncertainty estimation and convergence}}\\ \hdashline
Uncertainty floor
& $\sigma^{\min}$
& $0.2$
& Lower clipping bound applied to the empirical-Fisher-based
uncertainty estimate after temporal smoothing. \\

Stopping threshold
& $\sigma_{\tau}$
& $0.4$
& Threshold applied to the mean Fisher-based uncertainty for global
convergence checking. \\

\hdashline
\multicolumn{4}{l}{\textit{Numerical optimization}}\\ \hdashline
Adam configuration
&
$\bigl(\mathrm{lr},T_{\max},T_{\mathrm{pat}},
\epsilon_{\mathrm{num}}\bigr)$
&
$\bigl(0.01,2000,50,10^{-6}\bigr)$
&
Learning rate, maximum number of epochs, early-stopping patience,
and the numerical clipping constant used in each periodic global MAP
optimization. \\
\hline
\end{tabular}}
\caption{Core hyperparameters used for \textbf{human pairwise VAD annotation}, following the BPA formulation in the main text.}
\label{tab:human_bpa_hyperparameters}
\end{table*}

\subsection{Reliability-Weighted Committee Judgment Aggregation in E$^3$mo-Score}
\label{sec:committee_aggregation}

For each comparison, E$^3$mo-Score queries a committee of multimodal
judges and aggregates their ternary decisions through reliability-weighted voting. Model reliability is calibrated independently for each perspective-dimension subtask using a held-out screening set composed of predefined anchor combinations. For judge $m$, the screening score is
\begin{equation}
\begin{aligned}
S_m ={}&
0.40 A_m^{\mathrm{easy}}
+0.25 A_m^{\mathrm{medium}}
+0.10 A_m^{\mathrm{hard}}
+0.10 R_m^{\mathrm{sim}}\\
&+0.15 C_m^{\mathrm{flip}}
-0.50 E_m^{\mathrm{invalid}},
\end{aligned}
\label{eq:committee_screening_score}
\end{equation}
where the terms denote accuracy on the $3$ difficulty levels,
similar-pair recall, order-flip consistency, and invalid-output rate,
respectively. This score determines both a continuous reliability
coefficient
\begin{equation}
\nu_m
=
0.8+
\operatorname{clip}
\left(
0.8(S_m-0.5),\,0,\,0.4
\right)
\label{eq:committee_reliability}
\end{equation}
and a screening tier
$c_m\in\{\mathrm{low},\mathrm{mid},\mathrm{high}\}$. For low-tier
judges, $\nu_m$ is additionally capped at $0.95$.

The tier-based confidence multiplier is defined as
\begin{equation}
\chi(c_m)
=
\begin{cases}
\chi_{\mathrm{low}}=0.75,
& c_m=\mathrm{low},\\
\chi_{\mathrm{mid}}=1.00,
& c_m=\mathrm{mid},\\
\chi_{\mathrm{high}}=1.25,
& c_m=\mathrm{high}.
\end{cases}
\label{eq:committee_confidence_multiplier}
\end{equation}
Accordingly, the effective voting weight of judge $m$ is
$\omega_m=\nu_m\chi(c_m)$.

For pair $(i,j)$, each valid judge returns
$y_{ijm}\in\{+,0,-\}$, indicating that sample $i$ is higher, the two
samples are similar, or sample $j$ is higher, respectively. We encode
these outcomes using
$g(+)=1$, $g(0)=0$, and $g(-)=-1$, and compute the weighted committee
margin
\begin{equation}
M_{ij}
=
\sum_{m\in\mathcal{K}_{ij}}
\omega_m g(y_{ijm}),
\label{eq:committee_margin}
\end{equation}
where $\mathcal{K}_{ij}$ is the set of judges producing valid outputs.
The aggregated comparison is then
\begin{equation}
\widehat{y}_{ij}
=
\begin{cases}
+, & M_{ij}>\epsilon_{\mathrm{vote}},\\
0, & |M_{ij}|\leq\epsilon_{\mathrm{vote}},\\
-, & M_{ij}<-\epsilon_{\mathrm{vote}},
\end{cases}
\label{eq:committee_decision}
\end{equation}
where $\epsilon_{\mathrm{vote}}\geq0$ controls the committee-level
similarity region and is set to zero in the default configuration.

We additionally quantify unweighted committee disagreement as
\begin{equation}
D_{ij}^{\mathrm{com}}
=
1-
\frac{\max\{n_{ij}^{+},n_{ij}^{0},n_{ij}^{-}\}}
     {|\mathcal{K}_{ij}|},
\label{eq:committee_disagreement}
\end{equation}
where $n_{ij}^{+}$, $n_{ij}^{0}$, and $n_{ij}^{-}$ are the numbers of
valid votes for the three outcomes. The weighted margin captures the
direction and strength of the reliability-adjusted evidence, whereas
$D_{ij}^{\mathrm{com}}$ measures the categorical disagreement within
the committee. The aggregated label is subsequently treated as one
pairwise observation by the provisional TrueSkill update and the
periodic global BPA optimization.

\begin{table*}[t]
\centering
\renewcommand{\arraystretch}{1.08}
\setlength{\tabcolsep}{15pt}
\resizebox{\textwidth}{!}{
\begin{tabular}{l l c p{9.0cm}}
\hline
\textbf{Module} &
\textbf{Symbol} &
\textbf{Value} &
\textbf{Definition} \\
\hline

\multicolumn{4}{l}{\textit{Score initialization}}\\ \hdashline

Prior mapping range
& $[L,U]$
& $[2,8]$
& Output interval of the percentile-based mapping from preliminary
standardized VAD estimates to the initial scores of candidate
samples. \\

Fallback midpoint
& $\mu_{\mathrm{mid}}$
& $5.0$
& Neutral initial score used only when a valid preliminary VAD
estimate is unavailable. \\

Initial uncertainty
& $\sigma_{\mathrm{init}}$
& $1.333$
& Initial uncertainty assigned to candidate samples before online
pairwise updating. \\

Final score range
& $[\mu_{\min},\mu_{\max}]$
& $[1,9]$
& Feasible range imposed on candidate scores during periodic global
MAP optimization. \\

\hdashline
\multicolumn{4}{l}{\textit{Provisional TrueSkill updating}}\\ \hdashline

Performance-noise scale
& $\beta_{\mathrm{TS}}$
& $0.667$
& Standard deviation of the per-sample performance noise in the
provisional TrueSkill update. \\

Draw probability
& $p_{\mathrm{draw}}$
& $0.15$
& Prior probability of a similar outcome used to determine the
TrueSkill draw margin. \\

\hdashline
\multicolumn{4}{l}{\textit{Periodic global MAP optimization}}\\ \hdashline

Learning rate
& $\mathrm{lr}_{\mathrm{MAP}}$
& $0.05$
& Adam learning rate used during each periodic global MAP update. \\

Optimization steps
& $T_{\mathrm{MAP}}$
& $40$
& Number of gradient-based optimization steps performed in each
global update round. \\

Score-prior strength
& $\lambda_0$
& $0.05$
& Base coefficient of the adaptive quadratic prior centered at
$\mu_i^{(0)}$. \\

Comparison-count decay
& $\gamma$
& $0.25$
& Rate at which the score-prior weight decreases as the number of
comparisons $N_i$ involving sample $i$ increases. \\

\hdashline
\multicolumn{4}{l}{\textit{Uncertainty estimation and convergence}}\\ \hdashline
Uncertainty smoothing
& $\rho$
& $0.70$
& Temporal smoothing coefficient that combines the current
empirical-Fisher estimate with the uncertainty from the preceding
global update. \\
Uncertainty floor
& $\sigma^{\min}$
& $0.2$
& Lower clipping bound applied to the empirical-Fisher-based
uncertainty estimate after temporal smoothing. \\

Stopping threshold
& $\sigma_{\tau}$
& $0.25$
& Threshold applied to the mean Fisher-based uncertainty for global
convergence checking. \\

\hline
\end{tabular}
}
\caption{Core hyperparameters used by \textbf{E$^3$mo-Score}. The notation
follows the BPA formulation in the main text.}
\label{tab:e3mo_score_hyperparameters}
\end{table*}

\subsection{Probability-Based VAD Scoring Method}

Most evaluated MLLMs are not specifically trained to regress
continuous VAD scores. We therefore adapt the logit-pooling strategy
of Q-Bench~\cite{qbench} and the token-as-score formulation of
ArtiMuse~\cite{artimuse} to the $9$-point VAD scale. Unlike
three-level emotion-intensity pooling in EEmo-Bench~\cite{eemo-bench}, our formulation retains all
$9$ ordinal rating levels.

For each video $i$ and VAD dimension
$h\in\{\mathrm{V},\mathrm{A},\mathrm{D}\}$, the model is instructed to
assign an integer rating from 1 to 9 and to output only the rating.
The prompt is conditioned on the predefined affective perspective:
evoked-emotion questions concern the viewer's affective response,
whereas expressed-emotion questions concern the emotion conveyed by
the depicted subject. Let

\begin{equation}
    \mathcal{R}_s=\{1,2,\ldots,9\}
\end{equation}
denote the set of candidate ratings, and let
$\ell_{i,h}^{(r)}$ be the next-token logit associated with rating
$r\in\mathcal{R}_s$ at the designated score position. We normalize the
$9$ candidate logits as
\begin{equation}
    \rho_{i,h}^{(r)}
    =
    \frac{
        \exp\!\left(\ell_{i,h}^{(r)}\right)
    }{
        \displaystyle
        \sum_{u\in\mathcal{R}_s}
        \exp\!\left(\ell_{i,h}^{(u)}\right)
    }.
    \label{eq:vad_rating_probability}
\end{equation}
Here, $\rho_{i,h}^{(r)}$ is the probability of rating $r$ conditional
on the output belonging to the valid rating set $\mathcal{R}_s$. The
tokenizer-specific token corresponding to each numeral is used under
the same fixed response template for all samples.

The final VAD prediction is computed as the expectation of this
ordinal distribution:
\begin{equation}
    \widehat{r}_{i,h}
    =
    \sum_{r\in\mathcal{R}_s}
    r\,\rho_{i,h}^{(r)}.
    \label{eq:probability_vad_score}
\end{equation}
Consequently, $\widehat{r}_{i,h}\in[1,9]$ provides a continuous score
while preserving the model's relative confidence across adjacent
rating levels. This probability-weighted estimate avoids the
quantization introduced by directly decoding the highest-logit rating
token.

\begin{figure*}[t] 
    \centering
    \includegraphics[width=0.9\textwidth]{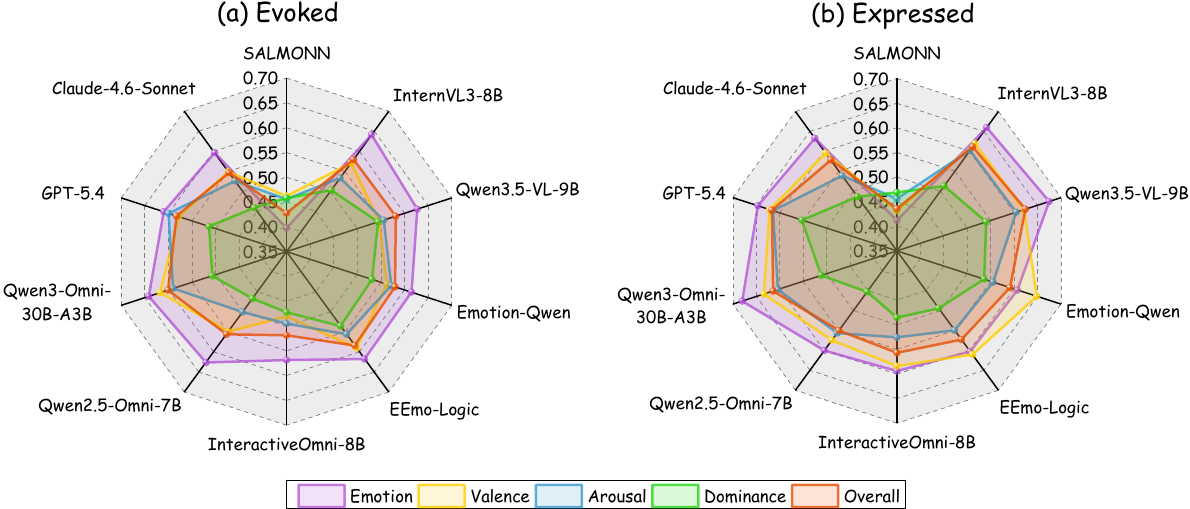}
    \caption{Performance comparison of $10$ representative MLLMs on perception QA tasks across diverse perceptual dimensions under both expressed and evoked perspectives.}
    \label{fig: perceptual radar}
\end{figure*}

\section{Implementation Details} \label{sec: implementation details}

\begin{table*}[]
\centering
\setlength{\tabcolsep}{3pt}  
\resizebox{\textwidth}{!}{
\begin{tabular}{lcccccccccc}
\hline
\multicolumn{1}{c|}{\textbf{Perspectives}}        & \multicolumn{5}{c|}{\textbf{Evoked}}                                                                                     & \multicolumn{5}{c}{\textbf{Expressed}}                                                              \\ \hline
\multicolumn{1}{c|}{\textbf{Dimensions}}          & \textit{Emotion↑} & \textit{Valence↑} & \textit{Arousal↑} & \textit{Dominance↑} & \multicolumn{1}{c|}{\textit{Overall↑}} & \textit{Emotion↑} & \textit{Valence↑} & \textit{Arousal↑} & \textit{Dominance↑} & \textit{Overall↑} \\ \hline
\multicolumn{1}{c|}{Random guess w/o recognition} & 35.98\%           & 43.02\%           & 43.05\%           & 43.00\%             & \multicolumn{1}{c|}{40.09\%}           & 35.98\%           & 43.03\%           & 43.01\%           & 43.01\%             & 40.13\%           \\ \hdashline
\multicolumn{11}{l}{\textit{Open-source audio-only MLLMs}}                                                                                                                                                                                                                         \\ \hdashline
\multicolumn{1}{l|}{Qwen-Audio~\cite{qwen-audio}}                   & 42.07\%           & 42.28\%           & 42.05\%           & 40.17\%             & \multicolumn{1}{c|}{41.14\%}           & 36.02\%           & 33.40\%           & 28.11\%           & 27.78\%             & 32.29\%           \\
\multicolumn{1}{l|}{SALMONN~\cite{salmonn}}                      & 39.73\%           & 46.22\%           & 45.57\%           & 45.83\%             & \multicolumn{1}{c|}{42.74\%}           & 41.19\%           & 43.16\%           & 45.60\%           & 46.88\%             & 43.57\%           \\ \hdashline
\multicolumn{11}{l}{\textit{Open-source video-only MLLMs}}                                                                                                                                                                                                                         \\ \hdashline
\multicolumn{1}{l|}{LLaVA-NeXT-Video-7B~\cite{llava-next}}          & 42.77\%           & 42.28\%           & 40.13\%           & 41.35\%             & \multicolumn{1}{c|}{41.88\%}           & 48.36\%           & 48.06\%           & 43.25\%           & 44.42\%             & 46.52\%           \\
\multicolumn{1}{l|}{VideoChat2~\cite{videochat2}}                   & 39.03\%           & 41.85\%           & 41.62\%           & 41.99\%             & \multicolumn{1}{c|}{40.66\%}           & 39.62\%           & 42.04\%           & 41.82\%           & 41.66\%             & 40.93\%           \\
\multicolumn{1}{l|}{Janus-Pro-7B~\cite{janus}}                 & 49.25\%           & 48.35\%           & 43.12\%           & 43.80\%             & \multicolumn{1}{c|}{46.82\%}           & 53.80\%           & 51.12\%           & 46.93\%           & 45.24\%             & 50.24\%           \\
\multicolumn{1}{l|}{InternVL3-8B~\cite{internvl3}}                 & \underline{64.11\%}     & 57.08\%           & 53.36\%           & 50.32\%             & \multicolumn{1}{c|}{57.97\%}           & 65.98\%           & 61.94\%           & 60.22\%           & 51.28\%             & 61.16\%           \\
\multicolumn{1}{l|}{Qwen3.5-VL-9B~\cite{qwen3.5}}                & 62.71\%           & 54.63\%           & 55.60\%           & \textbf{54.38\%}    & \multicolumn{1}{c|}{58.14\%}           & \underline{67.50\%}     & 62.14\%           & \underline{60.43\%}     & \underline{54.04\%}       & \textbf{62.41\%}  \\
\multicolumn{1}{l|}{AffectGPT~\cite{AffectGPT}}                    & 40.23\%           & 35.68\%           & 38.85\%           & 32.37\%             & \multicolumn{1}{c|}{37.55\%}           & 41.14\%           & 37.45\%           & 42.13\%           & 37.87\%             & 39.97\%           \\
\multicolumn{1}{l|}{Emotion-Qwen~\cite{emotion-qwen}}                 & 61.47\%           & 56.23\%           & 57.20\%           & 53.10\%             & \multicolumn{1}{c|}{57.99\%}           & 60.73\%           & \textbf{64.80\%}  & 55.52\%           & 53.63\%             & 59.11\%           \\
\multicolumn{1}{l|}{EEmo-Logic~\cite{eemologic}}                   & 61.76\%           & \underline{59.00\%}     & 55.60\%           & \underline{53.53\%}       & \multicolumn{1}{c|}{\underline{58.42\%}}     & 60.28\%           & 61.02\%           & 54.91\%           & 49.44\%             & 57.24\%           \\ \hdashline
\multicolumn{11}{l}{\textit{Open-source Omni-modal MLLMs}}                                                                                                                                                                                                                               \\ \hdashline
\multicolumn{1}{l|}{R1-Omni-0.5B~\cite{R1-Omni}}                 & 55.23\%           & 51.86\%           & 52.40\%           & 47.97\%             & \multicolumn{1}{c|}{52.62\%}           & 57.00\%           & 60.20\%           & 55.21\%           & 46.98\%             & 55.31\%           \\
\multicolumn{1}{l|}{InteractiveOmni-8B~\cite{interactiveomni}}           & 56.88\%           & 48.03\%           & 49.52\%           & 47.22\%             & \multicolumn{1}{c|}{51.85\%}           & 59.40\%           & 58.37\%           & 52.56\%           & 48.52\%             & 55.71\%           \\
\multicolumn{1}{l|}{Qwen2.5-Omni-7B~\cite{Qwen2.5-Omni}}              & 62.61\%           & 54.95\%           & 50.05\%           & 46.69\%             & \multicolumn{1}{c|}{55.59\%}           & 60.04\%           & 57.45\%           & 55.60\%           & 45.31\%             & 54.99\%           \\
\multicolumn{1}{l|}{Qwen3-Omni-30B-A3B~\cite{qwen3-omni}}           & \textbf{64.16\%}  & \textbf{61.77\%}  & \underline{59.02\%}     & 50.75\%             & \multicolumn{1}{c|}{\textbf{60.09\%}}  & \textbf{68.04\%}  & \underline{63.35\%}     & 60.37\%           & 51.15\%             & 61.42\%           \\ \hdashline
\multicolumn{11}{l}{\textit{Proprietary video-only MLLMs}}                                                                                                                                                                                                                          \\ \hdashline
\multicolumn{1}{l|}{GPT-5.4~\cite{gpt5.4}}                      & 61.02\%           & 57.83\%           & \textbf{59.98\%}  & 51.39\%             & \multicolumn{1}{c|}{58.32\%}           & 64.65\%           & 62.14\%           & \textbf{61.45\%}  & \textbf{55.27\%}    & \underline{61.69\%}     \\
\multicolumn{1}{l|}{Claude-4.6-Sonnet~\cite{claude_sonnet_46}}            & 59.62\%           & 54.95\%           & 52.61\%           & 45.94\%             & \multicolumn{1}{c|}{54.69\%}           & 63.28\%           & 59.59\%           & 53.89\%           & 48.62\%             & 57.82\%           \\ \hline
\end{tabular}
}
\caption{Performance evaluation across $4$ perceptual dimensions under each emotion perspective. The best and second-best results are \textbf{bolded} and \underline{underlined}, respectively.}
\label{tab: perceptual dimension}
\end{table*}

\subsection{Experiment Setup}

\subsubsection{Parameter Settings for Redundancy-Aware VAD Set Similarity $F^{\mathrm{VAD}}$.}
All distance-dependent parameters are determined from the empirical
distribution of pairwise distances between canonical emotions in the
standardized VAD space. Specifically, the redundancy threshold
$\tau_{\mathrm{dup}}$ is set to the $10$th percentile of this
distribution. The bandwidth
$\sigma_s$ is set to the corresponding $25$th percentile, providing a
data-adaptive scale for the base similarity. We set the similarity
sharpening coefficient to $\gamma=2.0$, which further suppresses
moderate-distance matches while preserving high-similarity pairs.
Finally, the cardinality-penalty exponent is set to $\beta=0.3$,
yielding a mild penalty for differences between the effective sizes of
the redundancy-reduced reference and prediction sets. Together, these
settings favor compact and affectively consistent matches without
imposing an overly strong penalty on reasonable variation in
open-vocabulary predictions.

\subsubsection{Parameter Settings for Human Pairwise Annotation.}
The key BPA hyperparameters used for human pairwise VAD annotation are summarized in Tab.~\ref{tab:human_bpa_hyperparameters}.

\subsubsection{Parameter Settings for E$^3$mo-Score.}

For evaluation, we partition the Golden Anchor Set into $14$ fixed anchors and $369$ candidate samples and conduct experiments separately across $6$ subtasks. Model inference is performed on $6$ NVIDIA RTX A6000 GPUs with $50$ GB of memory each. In total, $7{,}186$ pairwise comparisons are collected across the six subtasks. The key BPA hyperparameters are summarized in Tab.~\ref{tab:e3mo_score_hyperparameters}.

\subsection{Detailed Information of Evaluated MLLMs} \label{subsec: evaluated MLLMs}

\textbf{Qwen-Audio} \cite{qwen-audio} is an audio-language model that employs Whisper-large-v2~\cite{radford2023robust} as its audio encoder and Qwen-7B~\cite{qwen} as its language backbone. It adopts a single encoder to uniformly process diverse audio types, including speech, natural sounds, and music. It also supports over $30$ tasks, such as audio captioning, scene classification, emotion recognition, and audio question answering. Qwen-Audio-Chat further extends the base model through instruction fine-tuning for multi-turn dialogue.

\textbf{SALMONN} \cite{salmonn} is an audio-text multimodal model with a dual-encoder architecture, comprising Whisper as the speech encoder and BEATs~\cite{chen2022beats} as the audio encoder, with Vicuna (13B)~\cite{chiang2023vicuna} as the language backbone. It employs a window-level Q-Former~\cite{Q-former} as the connection module to convert variable-length encoder outputs into a variable number of text tokens, and uses low-rank adaptation (LoRA) as the cross-modal adaptor. The model is designed to jointly process speech, audio events, and music.

\textbf{LLaVA-NeXT-Video} \cite{llava-next} is a video understanding model built upon the LLaVA-NeXT architecture, with Qwen-1.5 as its language backbone. It adopts the AnyRes strategy to treat video frames as sequential image inputs, enabling a model trained solely on images to generalize directly to video tasks. DPO~\cite{DPO} training is further incorporated with video data to refine temporal understanding capabilities.

\textbf{VideoChat2} \cite{videochat2} is a video MLLM with UMT-L~\cite{li2023unmasked} as the visual encoder and Vicuna as the language backbone. It employs QFormer to compress visual features into a fixed number of query tokens, reducing computational overhead, while the QFormer is also instruction-aware, extracting task-relevant features conditioned on text instructions. LoRA is adopted as a lightweight adaptor between the visual encoder and the large language model (LLM). Training data covers both image and video modalities, with sources spanning conversation, visual question answering (VQA), reasoning, and classification tasks.

\textbf{Janus-Pro} \cite{janus} is a unified multimodal understanding and generation model built upon DeepSeek-LLM~\cite{bi2024deepseek}. It decouples visual encoding for understanding and generation: the understanding task employs a SigLIP~\cite{zhai2023sigmoid} encoder to extract semantic features, while the generation task uses a VQ tokenizer to convert images into discrete codes. The two types of features are mapped into the LLM space via respective adapters and processed jointly. This architecture mitigates representation conflicts between understanding and generation tasks through decoupled encoding pathways.

\textbf{InternVL3} \cite{internvl3} is a multimodal large language model with a ViT-MLP-LLM architecture, using InternViT as the visual encoder and Qwen2.5 as the language backbone. It adopts native multimodal pre-training, where all parameters are jointly optimized on both text-only and multimodal data during the pre-training phase. The model also incorporates Variable Visual Position Encoding (V2PE) to accommodate extended multimodal contexts, and employs Mixed Preference Optimization (MPO) as a post-training strategy.

\textbf{Qwen3.5-VL} \cite{qwen3.5} is a vision-language model from Alibaba Cloud's Qwen3.5 series. It features a hybrid architecture combining Gated DeltaNet (GDN) layers with standard attention layers, with SwiGLU activations and RMSNorm. The model supports both image and video understanding and adopts an Early Fusion architecture that enables end-to-end unified modeling of vision and language at the token level.

\textbf{AffectGPT} \cite{AffectGPT} is a multimodal large language model tailored for emotion understanding, leveraging CLIP ViT-L as the visual encoder, HUBERT-L as the audio encoder, and Qwen2.5 as the language backbone. Prior to the LLM's processing, the model incorporates a pre-fusion operation that explicitly integrates audio and visual features through either Q-Former or attention-based mechanisms. This architectural design extracts cross-modal interaction from the LLM, thereby reinforcing the integration of multimodal information.

\textbf{Emotion-Qwen} \cite{emotion-qwen} is a multimodal large language model tailored for emotion understanding, with CLIP ViT~\cite{clip} as the visual encoder and Qwen2.5 as the language backbone. A facial emotion capture (FEC) module detects faces and extracts emotion-relevant key frames from video inputs. A Hybrid Compressor based on the mixture of experts (MoE) paradigm is incorporated, comprising an emotion expert and a general expert with a gating network that dynamically routes inputs to balance emotion-specific and general-purpose processing.

\textbf{EEmo-Logic} \cite{eemologic} is a MLLM for image-evoked emotion assessment, built upon Qwen2.5-VL. It adopts a two-stage training framework with LoRA-based supervised fine-tuning, followed by group relative preference optimization (GRPO) with reward functions tailored for emotion ranking, VAD scoring, and discrete emotion classification. The model jointly supports both categorical (CES) and dimensional (DES) emotion representations for multi-granularity understanding.

\textbf{R1-Omni} \cite{R1-Omni} is an omni-modal large language model built upon HumanOmni~\cite{zhao2025humanomni}, supporting video frames and audio inputs for emotion recognition. It undergoes two-stage training: cold-start fine-tuning to initialize reasoning, followed by reinforcement learning with GRPO and reinforcement learning with verifiable rewards (RLVR), where rewards are based on accuracy and format compliance. The model outputs a reasoning process in \texttt{<think>} tags and the emotion prediction in \texttt{<answer>} tags.

\begin{table*}[]
\centering
\setlength{\tabcolsep}{7pt}        
\resizebox{\textwidth}{!}{
\begin{tabular}{l|cccc|cccc}
\hline
\multicolumn{1}{c|}{\textbf{Categories}} & \multicolumn{4}{c|}{\textbf{Evoked}}                                              & \multicolumn{4}{c}{\textbf{Expressed}}                                            \\ \hline
\multicolumn{1}{c|}{\textbf{Dimensions}} & \textbf{Valence}   & \textbf{Arousal}   & \textbf{Dominance} & \textbf{Overall}   & \textbf{Valence}   & \textbf{Arousal}   & \textbf{Dominance} & \textbf{Overall}   \\ \hline
InteractiveOmni-8B                          & 0.54~/~0.53          & 0.47~/~0.46          & 0.20~/~0.23          & 0.40~/~0.41          & 0.60~/~0.61          & 0.55~/~0.54          & 0.35~/~0.37          & 0.50~/~0.51          \\
Emotion-Qwen                             & 0.48~/~0.50          & 0.56~/~0.39          & 0.31~/~0.36          & 0.45~/~0.42          & \underline{0.66}~/~0.62          & 0.43~/~0.43          & \underline{0.46}~/~\underline{0.50}          & \underline{0.52}~/~\underline{0.52}          \\
EEmo-Logic                               & 0.52~/~0.52          & 0.39~/~0.39          & \underline{0.51}~/~\underline{0.53}          & 0.47~/~0.48          & 0.64~/~\underline{0.65}          & 0.39~/~0.39          & 0.42~/~0.44          & 0.48~/~0.49          \\
Qwen2.5-Omni-7B                             & \underline{0.66}~/~\underline{0.67}          & \underline{0.64}~/~\underline{0.63}          & 0.40~/~0.40          & \underline{0.57}~/~\underline{0.57}          & 0.53~/~0.55          & \underline{0.58}~/~\underline{0.59}          & 0.31~/~0.32          & 0.47~/~0.49          \\
Qwen3-Omni-30B-A3B                               & \textbf{0.72~/~0.72} & \textbf{0.73~/~0.72} & \textbf{0.55~/~0.50} & \textbf{0.67~/~0.65} & \textbf{0.81~/~0.76} & \textbf{0.71~/~0.69} & \textbf{0.61~/~0.62} & \textbf{0.71~/~0.69} \\ \hline
\end{tabular}
}
\caption{Overall performance of MLLMs on the assessment task across the entire E$^3$mo-Bench.}
\label{tab: assessment all}
\end{table*}

\textbf{InteractiveOmni} \cite{interactiveomni} is a unified omni-modal large language model that jointly processes image, video, audio, and text inputs, and generates both text and speech in an end-to-end manner for multi-turn audio-visual interaction. The architecture comprises InternViT as the visual encoder, Whisper as the audio encoder, Qwen3 as the language backbone, and Cosyvoice2~\cite{du2024cosyvoice} as the speech decoder. The model is designed for multi-turn dialogue and long-term memory in audio-visual interaction scenarios.

\textbf{Qwen2.5-Omni} \cite{Qwen2.5-Omni} is an end-to-end omni-modal model supporting text, image, audio, and video inputs with concurrent text and speech generation. It adopts a Thinker-Talker architecture, where the Thinker handles understanding and text generation and the Talker performs streaming speech synthesis. TMRoPE position encoding is introduced to align video and audio timestamps for real-time interaction.

\textbf{Qwen3-Omni} \cite{qwen3-omni} is an end-to-end omni-modal model with a MoE-based Thinker-Talker architecture. It replaces Whisper with a custom AuT audio encoder, adopts multi-codebook modeling and a lightweight ConvNet for streaming speech synthesis, and introduces TM-RoPE for multimodal timestamp alignment. The model supports $119$ text languages, $19$ speech understanding languages, and $10$ speech generation languages, with up to $40$ minutes of audio per instance.

\textbf{GPT-5.4} \cite{gpt5.4} is a frontier reasoning model from OpenAI that supports multimodal inputs including text, images, video, and audio, while generating both text and speech responses. It integrates the programming capabilities of GPT-5.3-Codex and is optimized for professional tasks such as spreadsheets, presentations, and documents. The model features native computer use for complex workflows across applications, supports up to $1M$ tokens of context, and offers a Thinking mode for reasoning transparency.

\textbf{Claude-4.6-Sonnet} \cite{claude_sonnet_46} is an omni-modal large language model developed by Anthropic, capable of jointly processing text, images, video, and audio inputs while generating both textual and image-based outputs. The model employs an adaptive reasoning mechanism that dynamically modulates the depth of inference in response to task complexity. It features native computer use capabilities and a $1$-million-token context window.

\begin{table*}[]
\resizebox{\textwidth}{!}{
\begin{tabular}{c|ccccccc}
\hline
Metrics~\textbackslash~Models & \multicolumn{1}{l}{InteractiveOmni} & \multicolumn{1}{l}{Emotion-Qwen} & \multicolumn{1}{l}{EEmo-Logic} & \multicolumn{1}{l}{Qwen2.5-Omni} & \multicolumn{1}{l}{Qwen3-Omni} & \begin{tabular}[c]{@{}c@{}}Committee\\ (Same Weight)\end{tabular} & \textbf{\begin{tabular}[c]{@{}c@{}}Committee\\ (Reliability-Weighted)\end{tabular}} \\ \hline
\textit{SRCC}                          & 0.66                                & 0.65                             & 0.68                           & 0.66                             & 0.70                           & 0.74                                                              & \textbf{0.76}                                                                       \\
\textit{PLCC}                         & 0.65                                & 0.61                             & 0.68                           & 0.65                             & 0.70                           & 0.75                                                              & \textbf{0.77}                                                                       \\ \hline
\end{tabular}
}
\caption{Ablation results of committee judgment aggregation strategies for E$^3$mo-Score on the evoked valence subtask.}
\label{tab: ablation 1}
\end{table*}

\begin{table}[]
\setlength{\tabcolsep}{10pt}        
\resizebox{\linewidth}{!}{
\begin{tabular}{c|ccccc}
\hline
Metrics~\textbackslash~Process & 20\% & 40\% & 60\% & 80\% & \textbf{100\%} \\ \hline
\textit{SRCC}                           & 0.68 & 0.74 & 0.75 & 0.76 & \textbf{0.76}  \\
\textit{PLCC}                           & 0.67 & 0.73 & 0.73 & 0.76 & \textbf{0.77}  \\ \hline
\end{tabular}
}
\caption{Ablation results of the BPA execution process for E$^3$mo-Score optimization on the evoked valence subtask.}
\label{tab: ablation 2}
\end{table}

\section{More Results Comparisons} \label{sec: additional results}

\subsection{Observation Across Perceptual Dimensions}

We categorize the Perception QA pairs into $4$ dimensions according to their target cues: \textbf{Emotion}, \textbf{Valence}, \textbf{Arousal}, and \textbf{Dominance}. Tab.~\ref{tab: perceptual dimension} reports the performance of all $16$ evaluated models, while Fig.~\ref{fig: perceptual radar} visualizes the comparative results of $10$ representative models. Overall, the models perform better on categorical emotion perception than on VAD polarity judgments, likely reflecting the broader coverage of categorical emotion tasks in existing MLLM training data. Valence and arousal yield comparable and relatively strong performance, whereas dominance remains substantially more challenging. Although the random baselines are similar across dimensions, the best dominance result is more than $5\%$ lower than the corresponding valence and arousal results, indicating that dominance requires a deeper understanding of agency, control, and contextual affect. GPT-5.4 achieves the strongest overall performance on the less extensively studied arousal and dominance dimensions, suggesting that broad multimodal training remains beneficial when task-specific supervision is limited. Notably, the medium-scale, open-source, video-only Qwen3.5-VL achieves $54.38\%$ and $54.04\%$ on evoked- and expressed-emotion dominance, ranking first and second, respectively. This result highlights its potential as a backbone for future emotion-specialized models.

\subsection{Observation on Assessment Task}

The main paper evaluates $5$ open-source models on the Golden Anchor Set for comparison with E$^3$mo-Score. To provide a more comprehensive assessment, we further evaluate their VAD predictions on the \textbf{full E$^3$mo-Bench}, with results reported in Tab.~\ref{tab: assessment all}. Qwen3-Omni-30B-A3B achieves the best performance across all $6$ subtasks, consistent with the main-paper findings and suggesting that its larger scale and omni-modal design better support the integration of complementary audio-visual cues for fine-grained affective intensity assessment.
The remaining models exhibit clear perspective-dependent differences despite having comparable parameter scales. For evoked emotion, Qwen2.5-Omni-7B achieves the second-best overall performance, indicating the importance of jointly modeling auditory and visual information. For expressed emotion, Emotion-Qwen ranks second overall, likely benefiting from its pretraining and fine-tuning on large-scale facial-expression datasets. This result further suggests that expressed-emotion assessment depends more strongly on visual behavioral cues. Across all models, valence and arousal are assessed more reliably than dominance, whose abstract notion of control and agency remains difficult to model. Improving dominance assessment therefore remains an important direction toward more comprehensive affect understanding.

\subsection{Additional Ablation Studies}

Using the evoked-valence subtask as a representative case, we conduct a more detailed ablation analysis of E$^3$mo-Score as follows:

\subsubsection{Ablation Study on Committee Judgment Aggregation.}

As shown in Tab.~\ref{tab: ablation 1}, we first compare BPA optimization in E$^3$mo-Score using pairwise judgments from individual models and the full committee. Although Qwen3-Omni outperforms the other four individual models, its single-model results remain inferior to those of the five-model committee, demonstrating the benefit of aggregating complementary judgments. We further remove the confidence weights derived from model-reliability calibration and assign equal weights to all committee members. The resulting performance is slightly lower than that of the full method, indicating that capability-aware weighting helps preserve complementary evidence while reducing the influence of noisier judgments from weaker models.

\subsubsection{Ablation Study on Execution Process.}

To characterize the optimization trajectory of E$^3$mo-Score, we partition the ordered comparison records into five cumulative checkpoints at $20\%$ intervals. At each checkpoint, BPA re-estimates all candidate scores using the comparisons collected up to that point, and the resulting scores are correlated with the reference MOS ratings. As reported in Tab.~\ref{tab: ablation 2}, the inferred scores already exhibit substantial agreement with human ratings at the $20\%$ checkpoint and improve progressively as additional comparisons are incorporated, supporting the effectiveness of iterative BPA refinement. The gains become smaller beyond approximately $40\%$ of the comparison budget, indicating diminishing returns. In human annotation, this trend enables the stopping threshold $\sigma_{\tau}$ to be selected according to the desired trade-off between annotation cost and score accuracy. For model-based scoring, where additional judgments incur substantially lower marginal cost, a larger comparison budget can be used to obtain further, albeit progressively smaller, improvements.

\section{Additional Qualitative Results} \label{sec: qualitative results}

\begin{figure*}[t] 
    \centering
    \includegraphics[width=\textwidth]{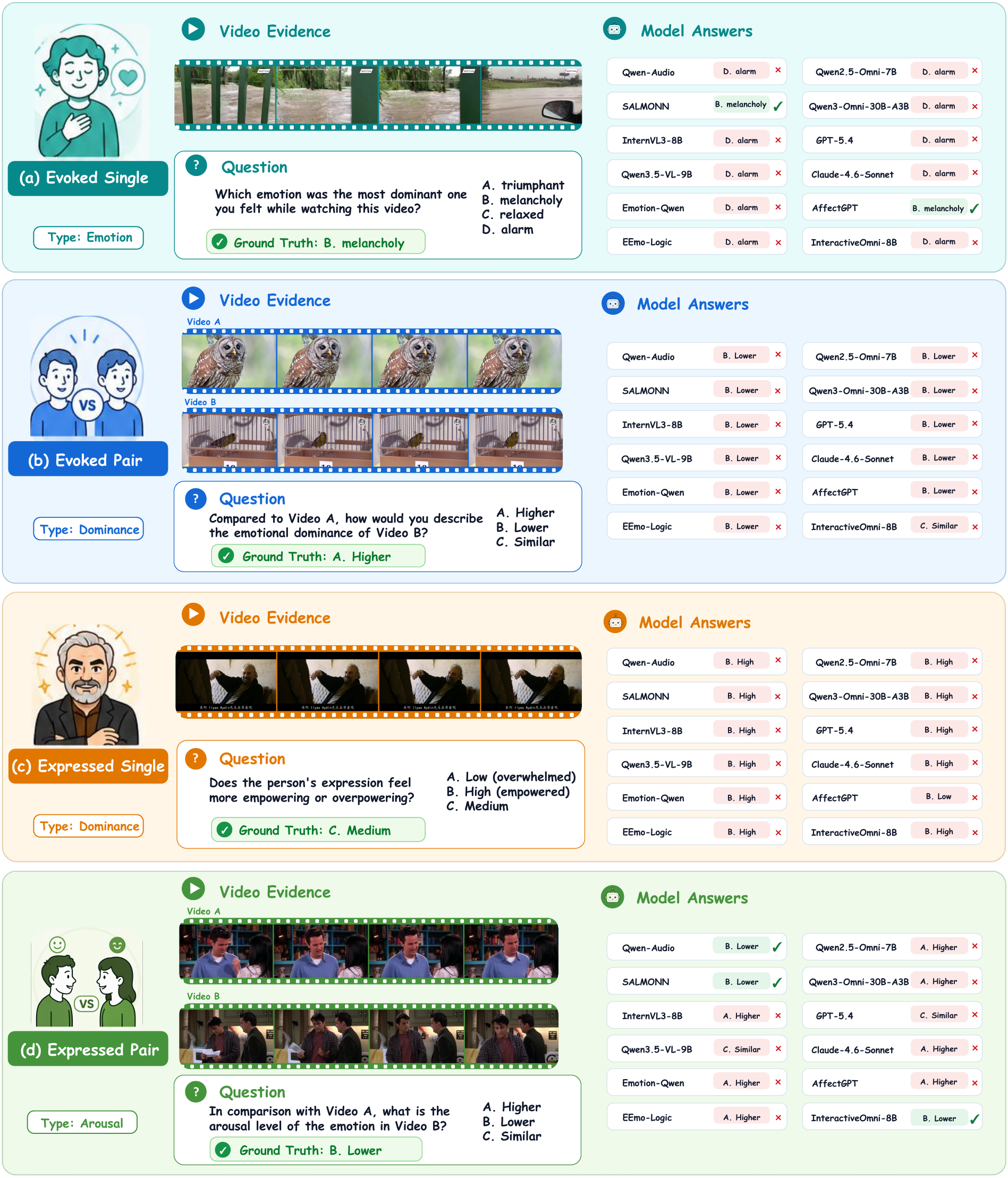}
    \caption{Failure case analysis of the QA results across four perceptual dimensions. The analysis encompasses both expressed and evoked emotions under single-video and pairwise evaluation scenarios.}
    \label{fig: QA failure case}
\end{figure*}

\begin{figure*}[t] 
    \centering
    \includegraphics[width=0.95\textwidth]{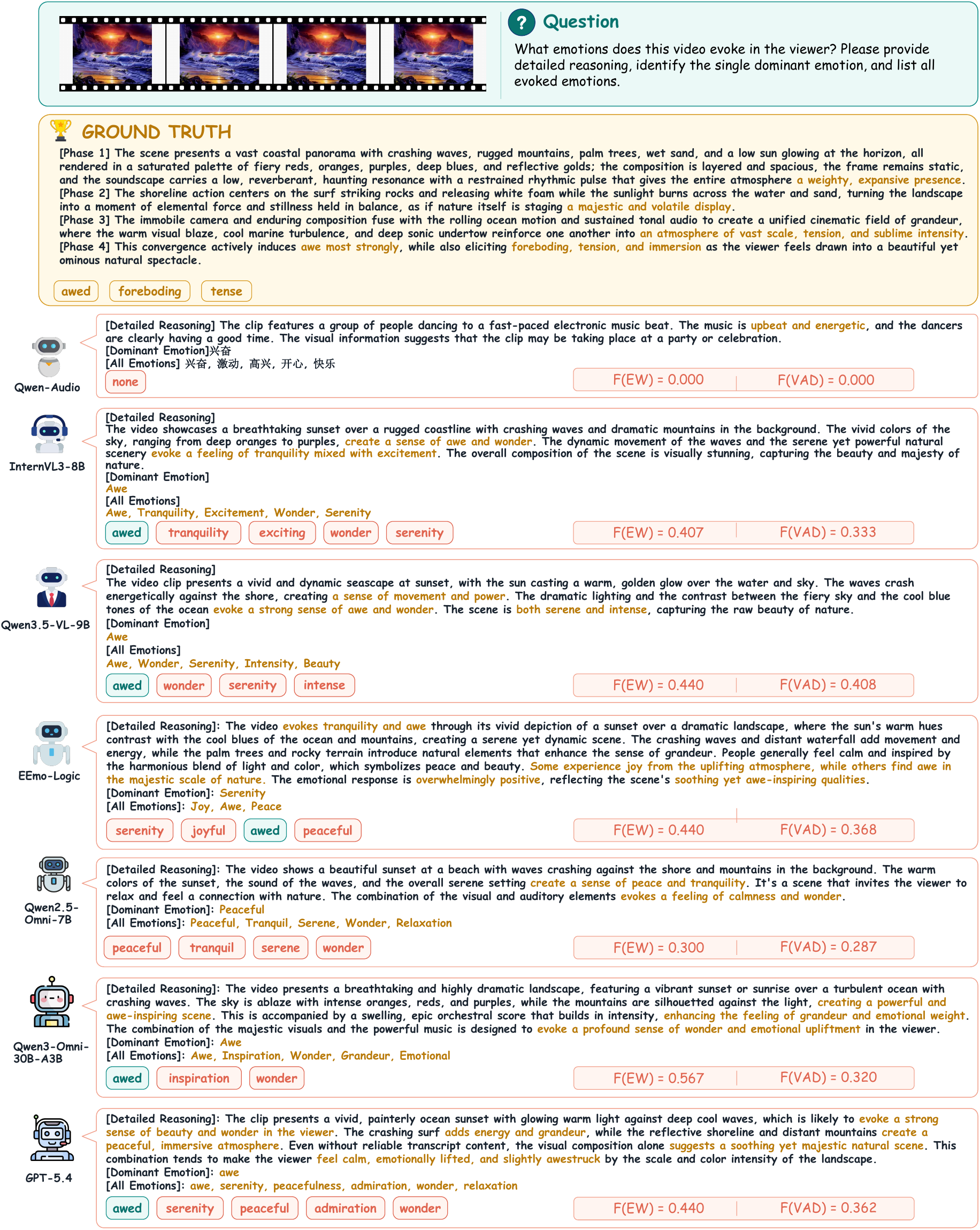}
    \caption{Visualization of a challenging evoked emotion case in OV emotion recognition and reasoning.}
    \label{fig: OV evoked case}
\end{figure*}

\begin{figure*}[t] 
    \centering
    \includegraphics[width=0.95\textwidth]{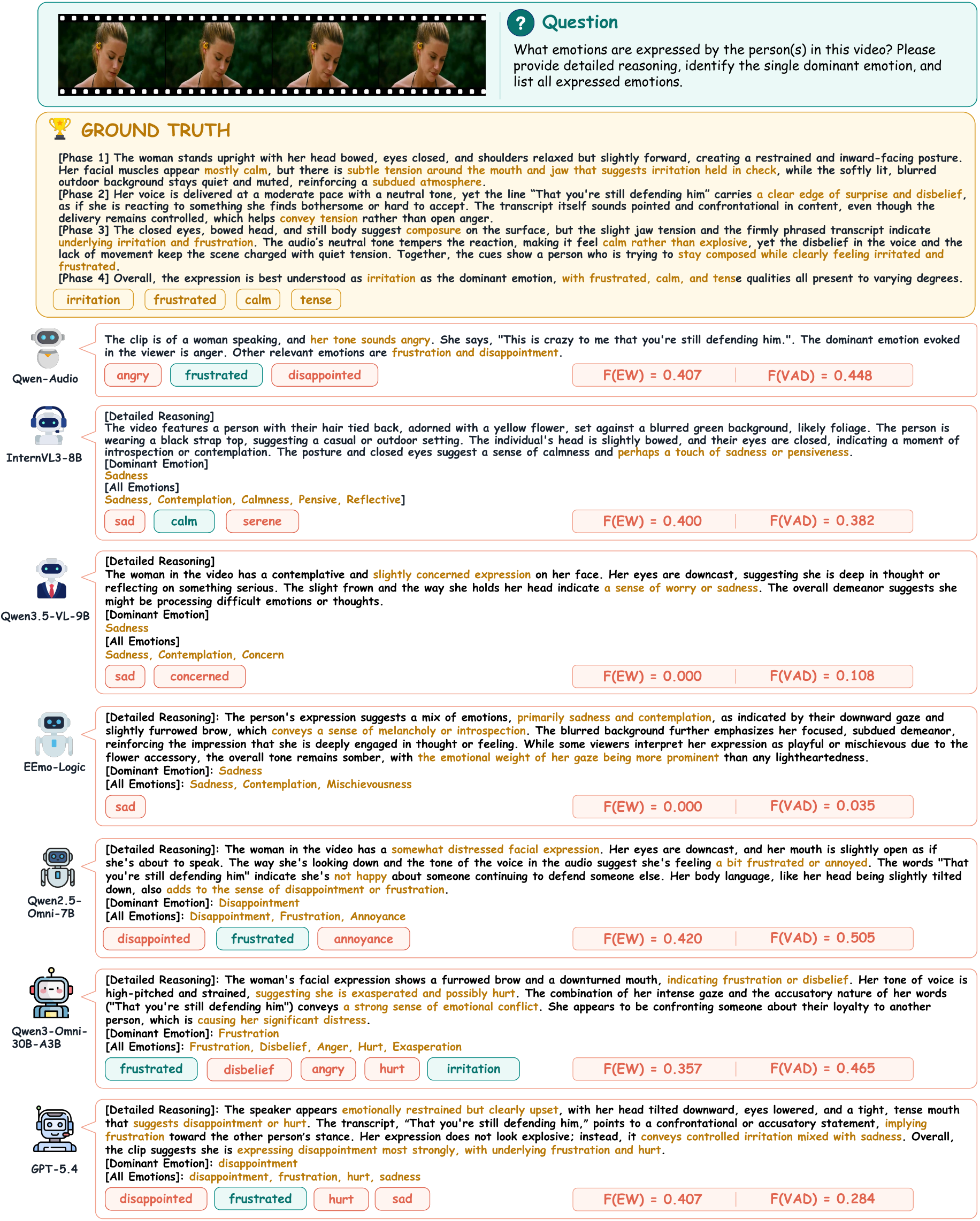}
    \caption{Visualization of a challenging expressed emotion case in OV emotion recognition and reasoning.}
    \label{fig: OV expressed case}
\end{figure*}

We present qualitative failure cases for the Perception task across $4$ perceptual dimensions, covering both expressed and evoked emotions under single-video and pairwise evaluation settings, as shown in Fig.~\ref{fig: QA failure case}. We further analyze challenging cases of OV emotion recognition and reasoning for both affective perspectives in Figs.~\ref{fig: OV evoked case} and \ref{fig: OV expressed case}.

\section{Limitations} \label{sec: limitations}

\subsection{Perspective-Conditioned Annotation}

Expressed and evoked emotions are not mutually exclusive and may coexist within the same audio-visual content. E$^3$mo-Bench therefore adopts a perspective-conditioned evaluation design, in which each video is assigned a predefined primary perspective that specifies the affective target the model is expected to analyze, rather than implying that the video contains only one type of emotion. This sample-level perspective specification reduces target ambiguity and supports scalable annotation, while the benchmark as a whole covers both expressed and evoked emotion understanding. Nevertheless, the current design does not provide paired annotations for both perspectives on every video and therefore cannot directly characterize their within-video relationship. Future work may introduce a dual-perspective subset in which the same content is independently annotated under both affective targets.

\subsection{BPA Configuration and Sensitivity}
Although BPA is grounded in established theories of pairwise
comparison and Bayesian inference, its concrete dispatch policy,
optimization schedule, and hyperparameter configuration are specific
to this work. Parameters such as the proportions of the dispatch
strategies, score-gap thresholds, update frequency, and stopping
criteria are selected according to their theoretical roles and
preliminary validation, rather than through an exhaustive
hyperparameter search. Nevertheless, the anchor-recovery results,
convergence analyses, and component ablations show that the resulting
configuration reliably transforms sparse comparisons into
anchor-aligned VAD scores. These results suggest that the observed
performance primarily arises from the overall BPA design rather than
from narrowly tuned parameter values, although they do not establish
complete insensitivity to the selected configuration. Future work will
conduct broader sensitivity analyses and investigate adaptive or
data-driven parameter selection to further improve the reliability and
efficiency of comparison-to-score inference.

\clearpage
\bibliography{supp_reference}
